\documentclass[11pt]{article}

\usepackage[table,xcdraw]{xcolor}
\definecolor{NLcol}{HTML}{C04F15}
\definecolor{FOLcol}{HTML}{3B7D23}
\definecolor{MIXcol}{HTML}{1E77B4}
\definecolor{Finalcol}{HTML}{FF800E}
\definecolor{CMTgreen}{HTML}{228B22}

\usepackage[final]{acl}

\hypersetup{
    colorlinks=true,
    urlcolor=magenta,
    linkcolor=red
}

\usepackage{siunitx}
\usepackage{url}
\usepackage{amsthm}
\usepackage{booktabs}
\usepackage{multirow}
\usepackage{pifont}
\usepackage{listings}
\usepackage{xcolor}
\definecolor{codepurple}{rgb}{0.58,0,0.82}
\definecolor{codegray}{rgb}{0.5,0.5,0.5}
\definecolor{codeblue}{rgb}{0.0,0.2,0.6}
\definecolor{codered}{rgb}{0.8,0.0,0.0}
\definecolor{backcolour}{rgb}{0.98,0.98,0.98}

\usepackage{algorithm}
\usepackage{algorithmicx}
\usepackage{algpseudocode}
\usepackage{makecell}
\usepackage[most]{tcolorbox}

\usepackage{enumitem}
\usepackage{amssymb} 

\usepackage{algpseudocode}
\usepackage{subcaption}
\usepackage{colortbl}

\usepackage{amsthm}

\theoremstyle{definition}
\newtheorem{definition}{Definition}   

\theoremstyle{plain}

\usepackage{times}
\usepackage{latexsym}

\usepackage[T1]{fontenc}

\usepackage[utf8]{inputenc}

\usepackage{microtype}

\usepackage{inconsolata}

\usepackage{graphicx}

\title{Logical Phase Transitions: Understanding Collapse in LLM Logical Reasoning}

\author{
 \textbf{Xinglang Zhang\color{red}{\footnotemark[1]}}, 
 \textbf{Yunyao Zhang\thanks{Equal contribution. Order determined by coin flip.}}, 
 \textbf{Zeliang Chen},
 \textbf{Junqing Yu},
 \textbf{Wei Yang},
 \textbf{Zikai Song\thanks{Corresponding author}} 
\\
 Huazhong University of Science and Technology
\\
 \small{
   \{normanspark, ikostar, skyesong\}@hust.edu.cn
 }
}

\begin{document}

\maketitle

\begin{abstract}
%

Symbolic logical reasoning is a critical yet underexplored capability of large language models (LLMs), providing reliable and verifiable decision-making in high-stakes domains such as mathematical reasoning and legal judgment.
In this study, we present a systematic analysis of logical reasoning under controlled increases in logical complexity, and reveal a previously unrecognized phenomenon, which we term \textbf{Logical Phase Transitions}: rather than degrading smoothly, logical reasoning performance remains stable within a regime but collapses abruptly beyond a critical logical depth, mirroring physical phase transitions such as water freezing beyond a critical temperature threshold.
Building on this insight, we propose \textbf{Neuro-Symbolic Curriculum Tuning}, a principled framework that adaptively aligns natural language with logical symbols to establish a shared representation, and reshapes training dynamics around phase-transition boundaries to progressively strengthen reasoning at increasing logical depths. 
Experiments on five benchmarks show that our approach effectively mitigates logical reasoning collapse at high complexity, yielding average accuracy gains of +1.26 in naive prompting and +3.95 in CoT, while improving generalization to unseen logical compositions. 
Code and data are available at: \url{https://github.com/AI4SS/Logical-Phase-Transitions}.

\end{abstract}

\section{Introduction}

\begin{figure}
  \includegraphics[width=\linewidth]{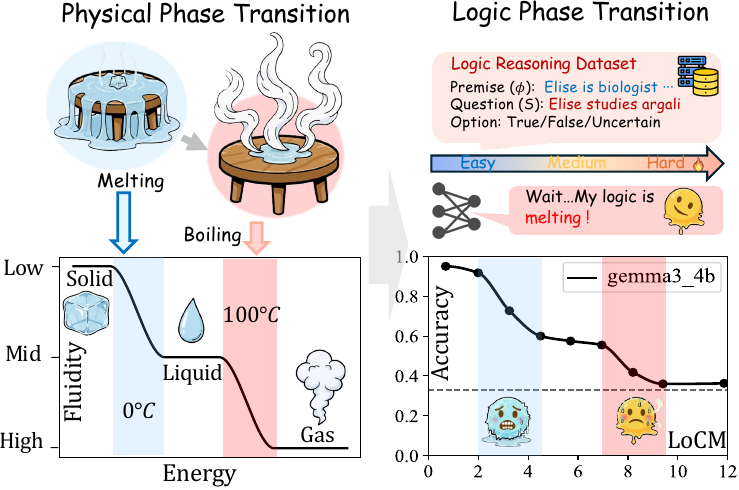}
  \caption{
    Overview of our work. 
    \textbf{(Left)} Physical phase transitions: as temperature increases, matter undergoes abrupt state changes (solid $\rightarrow$ liquid at $0^\circ$C, liquid $\rightarrow$ gas at $100^\circ$C).
    \textbf{(Right)} Logical phase transitions: as the Logical Complexity Metric (LoCM) increases, LLM reasoning accuracy drops sharply, revealing a phase-transition-like behavior in logical reasoning.
  }
  \label{fig:teaser}
\end{figure}

Symbolic logical reasoning refers to the ability to draw correct conclusions by applying explicit logical rules to structured premises~\citep{formal-logic-2003introduction}. It is a foundational aspect of human cognition and a core capability expected of large language models (LLMs)~\citep{AGI-2020asymptotically, Attention-2017attention}. It supports a wide spectrum of tasks, including commonsense inference~\citep{Commonsense-reasoning-2024-candle, liu2026chartverse}, mathematical proof~\citep{Mathematical-reasoning-2023mathcoder, Geometric-reasoning-2024deep,an2025amo}, and philosophical analysis~\citep{philosophical-thinking-2013critical, lin2026scientific}. 

Recent studies~\citep{Symbolic-COT-2024faithful,Baseline-Aristotle2024xu, LogicLM-2023} indicate that current LLMs perform well on logical reasoning tasks in simple scenarios, yet experience substantial degradation as reasoning becomes more challenging. While this gap is widely observed, how logical depth shapes reasoning ability in LLMs remains poorly understood, as there is still no clear characterization of reasoning behavior across increasing levels of logical complexity.


Motivated by this limitation, we analyze LLM logical reasoning from the perspective of logical complexity and uncover a previously unrecognized collapse in reasoning behavior. 
\textbf{(1)} \textit{Measuring}. We begin by proposing the \textbf{Logical Complexity Metric (LoCM)}, a principled measure that quantifies logical difficulty in terms of symbolic structure and compositional depth. To support this complexity-aware analysis, we further construct a logic-enhanced dataset that provides explicit first-order-logic (FOL) representations of propositions, premise sets, and reasoning chains, enabling fine-grained characterization of logical dependencies and structures. 
\textbf{(2)} \textit{Discovery}. Based on this measuring, we reveal a previously unrecognized phenomenon, termed \textbf{Logical Phase Transitions (LPTs)}: rather than degrading smoothly, reasoning performance remains stable within specific LoCM regimes and then collapses abruptly at critical thresholds, forming multiple LPTs across the whole LoCM range. This behavior mirrors successive phase transitions in physical systems, such as the transitions from ice to water and from water to vapor (solid–liquid–gas) as temperature increases, as illustrated in Figure~\ref{fig:teaser}.

Building on this collapse phenomenon LPTs, we propose \textbf{Neuro-Symbolic Curriculum Tuning}, a principled framework that explicitly aligns neural representations with symbolic structures and systematically reshapes the training dynamics around these phase-transition boundaries, to mitigate reasoning collapse under increasing logical complexity. Specifically, our framework consists of two core components: 
(1) \textit{\textbf{Adaptive Neuro-Symbolic Alignment}} establishes a shared representational space between natural language and logical symbols, enabling consistent reasoning across symbolic structures. 
(2) \textit{\textbf{Complexity-Aware Curriculum Optimization}} organizes training into successive stages of progressively increasing logical complexity, allowing models to gradually adapt to deeper reasoning structures, particularly near critical transition regions where collapse is likely to occur.

In summary, our contributions are:
\begin{itemize}[leftmargin=10pt, topsep=2pt, itemsep=0pt]
    \item We introduce the \textbf{Logical Complexity Metric} together with a logic-enhanced dataset, enabling fine-grained analysis of logical reasoning, and uncover a previously unrecognized collapse phenomenon, \textbf{Logical Phase Transitions}, characterized by performance collapse beyond critical complexity thresholds.
    \item We propose \textbf{Neuro-Symbolic Curriculum Tuning}, a principled framework that aligns neural and symbolic representations and restructures training around phase-transition boundaries, thereby mitigating reasoning collapse as logical complexity increases.
\end{itemize}

\begin{figure*}[t]
\centering
\includegraphics[width=\linewidth, height=0.45\linewidth]{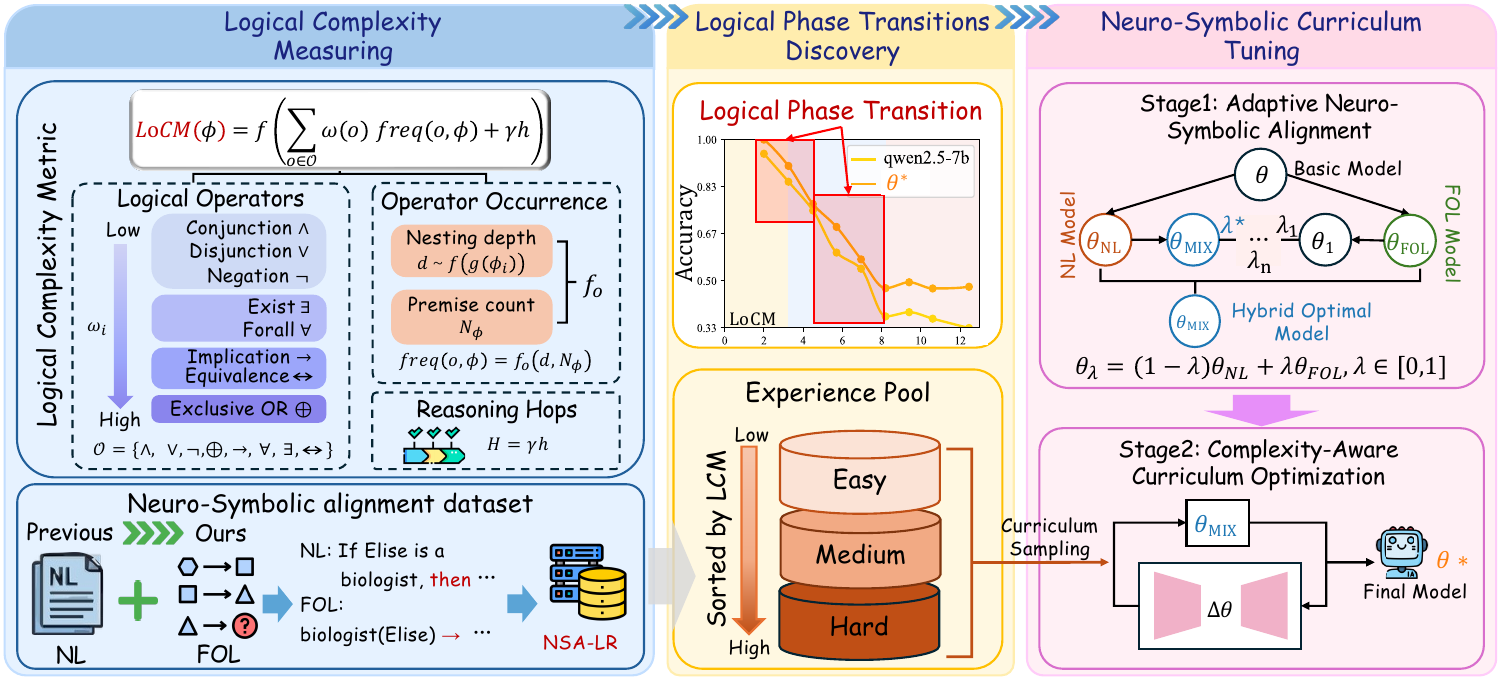}
\caption{
\textbf{Overview.}
\textbf{Logical Complexity Measuring (left):} We construct a Neuro-Symbolic alignment dataset (NSA-LR) and define the Logical Complexity Metric (LoCM), which quantifies reasoning difficulty via logical operators, nesting depth $d$, premise count $N_{\phi}$, and reasoning hops $H$.
\textbf{Logical Phase Transitions Discovery (center):} LoCM reveals a phase transition where accuracy collapses to random guessing beyond a critical complexity threshold and partitions samples into Easy, Medium, and Hard pools.
\textbf{Neuro-Symbolic Curriculum Tuning (right):} Stage 1 trains a mixed-semantics model \textcolor[HTML]{1E77B4}{$\theta_{\text{MIX}}$} and stage 2 applies curriculum optimization over the experience pool to obtain the final model \textcolor[HTML]{FF800E}{$\theta^{*}$}, mitigating reasoning collapse at high LoCM.
}
\label{fig2:pipline}
\end{figure*}

\section{Related Work}
A full discussion of related work appears in Appendix~\ref{app:related_work}; we highlight key directions here.

\noindent\textbf{LLM Logical Reasoning.}  
Prior work spans three main directions:  
(1) \textbf{Linear Reasoning (LR)} methods such as Naive Prompting and Chain-of-Thought~\citep{COT-2022chain};  
(2) \textbf{Aggregative Reasoning (AR)} approaches~\citep{CLOVER-ICLR-2024divide, Cumulative-reasoning-2023-TsinghuIIIS, TOT-2023tree, Determlr-ACL-2024-RenDaGaoling} that combine multiple reasoning trajectories; and  
(3) \textbf{Symbolic Reasoning (SR)} frameworks~\citep{NL2FOL-2023harnessing, Baseline-Aristotle2024xu, Symbolic-COT-2024faithful, LogicAgent-2025-ours} that integrate LLMs~\citep{Gpt4-2023gpt, DeepseekR1-2025deepseek, LLM-survey-2023survey} with explicit logic modules~\citep{LogicLM-2023}.  
While these methods improve surface performance through guided or modular reasoning, they offer limited insight into how reasoning behavior changes with increasing logical complexity.

\noindent\textbf{Logical Reasoning Capacity Analysis.}  
Recent work analyzes LLM reasoning fidelity, compositionality, and failure modes~\citep{Reasoning-survey-2022towards, multi-step-2024exploring,fang2025dualvla}. 
Studies such as InfoQA~\citep{InfoQA-ICLR2025-XiuyingChen-MBZUAI}, CoT-Valve~\citep{CoT-Valve-ACL2025-NUS}, and symbolic Monte Carlo supervision~\citep{Sym-MC-EMNLP2025-Sheffield} investigate capacity limits, chain-length controllability, and the role of symbolic trajectories, while \textit{Apple}~\citep{illusion-of-thinking-Nips2025-Apple} shows that frontier LMMs suffer sharp reasoning collapses in tasks such as Tower of Hanoi as complexity increases. 
However, their setting focuses on procedural execution in structured puzzles, whereas we study symbolic logical reasoning under explicit propositional/FOL structure; the two works therefore differ in complexity definition, evaluation target, and intervention scope. 
More broadly, these analyses rely on task-specific or coarse difficulty proxies and do not provide a principled framework for quantifying logical complexity itself or characterizing collapse behaviors.


\begin{table}[t]
\centering
\scriptsize
\setlength{\tabcolsep}{3pt}
\renewcommand{\arraystretch}{0.92}
\caption{
Comparison of logical reasoning datasets. Complete metric definitions and full dataset comparisons are provided in the Appendix~\ref{appendix:datasets_and_baselines}.
}
\label{tab1:dataset}
\newcommand{\cmark}{\textcolor{green!60!black}{\ding{51}}}
\newcommand{\xmark}{\textcolor{red!70!black}{\ding{55}}}

\begin{tabular}{lcccccc}
\toprule
\makecell{\textbf{Dataset}} &
\makecell{\textbf{Creation}} &
\makecell{\textbf{Scal.}} &
\makecell{\textbf{Natural}\\\textbf{Language}} &
\makecell{\textbf{Symbolic}\\\textbf{Rep.}} &
\makecell{\textbf{Faithful}\\\textbf{Chains}} &
\makecell{\textbf{Full}\\\textbf{FOL}} \\
\midrule
ProntoQA       & Synthetic & \cmark & \xmark & \cmark & \cmark & \xmark \\
ProofWriter    & Synthetic & \cmark & \xmark & \xmark & \cmark & \xmark \\
FOLIO          & Manual    & \xmark & \cmark & \cmark & \xmark & \xmark \\
ProverQA       & Synthetic & \cmark & \cmark & \cmark & \cmark & \xmark \\
\midrule
NSA-LR(Ours)          & Synthetic & \cmark & \cmark & \cmark & \cmark & \cmark \\
\bottomrule
\end{tabular}
\end{table}

\section{Methodology}
We describe our method along three aspects:
\S~\ref{sec3.1:measuring} introduces logical complexity measuring, including logical complexity metric and the Neuro-Symbolic alignment dataset;
\S~\ref{sec3.2:discovery} presents the discovery of logical phase transitions;
\S~\ref{sec3.3:tuning} presents Neuro-Symbolic Curriculum Tuning.
The overall framework is illustrated in Figure~\ref{fig2:pipline}.

\subsection{Logical Complexity Measuring}
\label{sec3.1:measuring}
To enable controlled analysis under increasing logical complexity, we introduce LoCM to quantify symbolic difficulty and construct a neuro-symbolic alignment dataset that provides the structured representations required to compute this metric.

\subsubsection*{Logical Complexity Metric}
Given an input proposition \(Q\), a model must integrate several symbolic factors such as the number of premises, the influence of logical operators, the depth of nested reasoning, and the overall length of the reasoning chain to determine its truth value. Existing complexity estimates are typically coarse and rely mainly on hop counts. To provide a more precise measure, we introduce the LoCM, which assigns each sample a scalar score that captures its logic difficulty.

\begin{definition}[LoCM]
\label{def:eic}
For a reasoning instance $\phi$ expressed in FOL, let 
$\mathcal{P}=\{p_1,\dots,p_{N_{\phi}}\}$ denote the set of premises, where 
$N_{\phi}=|\mathcal{P}|$. 
Let $
\mathcal{O}
=\{\land,\ \lor,\ \lnot,\ \oplus,\ \rightarrow,\ \leftrightarrow,\ \forall,\ \exists\} $
denote the set of logical operators, including Boolean connectives and quantifiers.
For each operator $o\in\mathcal{O}$, let  
$\mathrm{freq}(o,\phi)=f_o(d, N_{\phi})$ denote its occurrence count in $\phi$, 
where $d$ denotes the maximum syntactic nesting depth at which $o$ appears. 
Let $h$ denote the number of reasoning hops in the corresponding reasoning chain.  
The LoCM is defined as:
\begin{equation}\small
\mathrm{LoCM}(\phi)
= f\!\left(
        \sum_{o\in\mathcal{O}} \omega(o)\,\mathrm{freq}(o,\phi)
        + \gamma h(\phi)
    \right)
\label{eq:LoCM}
\end{equation}
where $\omega(o)$ assigns a symbolic-complexity weight, and $f(\cdot)$ is a monotonic transformation function used to stabilize scale and better fit empirical correlations. 
The metric $\mathrm{LoCM}(\phi)$ yields a single scalar score that quantifies the logical difficulty of a given reasoning instance.

\end{definition}

\begin{figure*}[t]
    \centering
    \includegraphics[width=\linewidth]{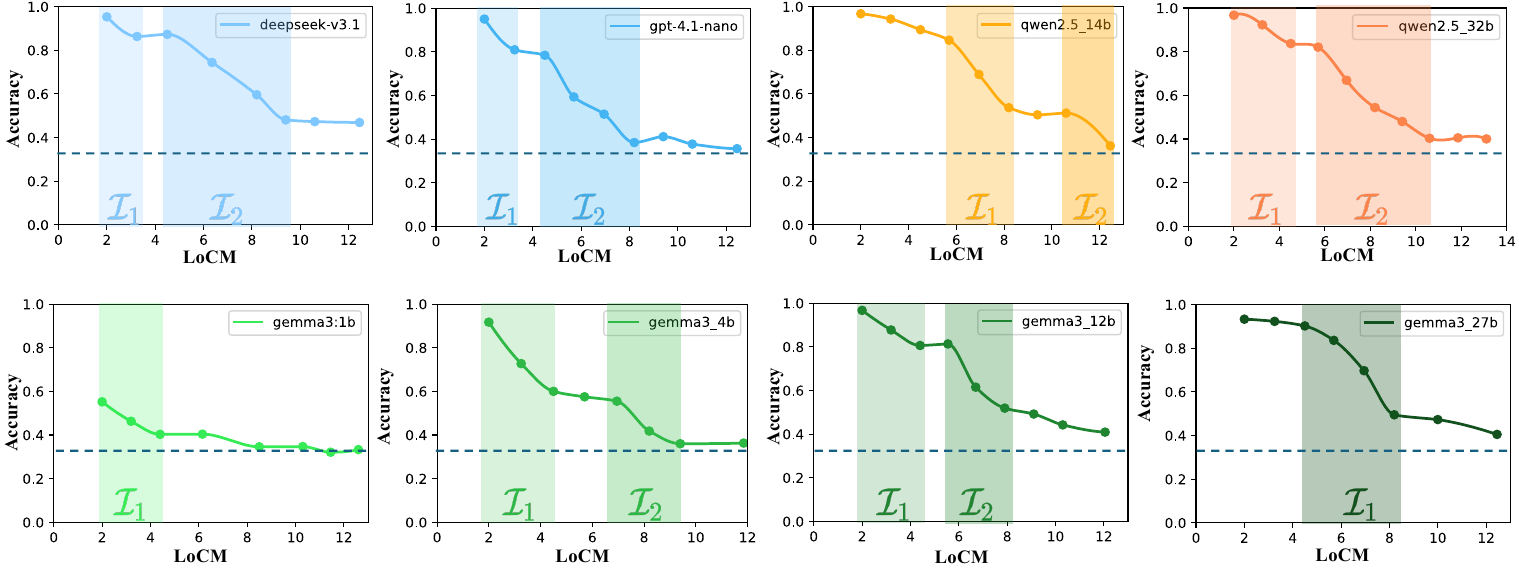}
    \caption{
    Logical phase-transition curves across different models. 
    Shaded regions denote the identified transition intervals, where accuracy drops sharply as LoCM increases. 
    The dashed line marks the $1/3$ baseline corresponding to random guessing. 
    Complete numerical results are provided in the Appendix~\ref{app:full_results}.
    }
    \label{fig3:LPT}
\end{figure*}

\subsubsection*{Neuro-Symbolic Alignment Dataset}
Building on the data-construction principles~\citep{Benchmark-ProverQA-ICLR-2025large}, we construct a \textbf{Neuro-Symbolic Alignment Dataset for Logical Reasoning (NSA-LR)} that provides paired natural language (NL) and FOL representations for every sample. All NL propositions, premises, and reasoning steps are translated into explicit predicates, quantifiers, connectives, and multi-step reasoning chains, following rules in Appendix Table~\ref{tab_app:fol_translation_rules}. And each statement is independently translated by GPT-5 and Qwen3-Max~\cite{Qwen3-2025qwen3}; matching outputs undergo CFG validation, while mismatches are manually adjudicated. A comparison with existing benchmarks, including ProofWriter~\citep{Benchmark-ProofWriter-2020proofwriter}, ProntoQA~\citep{Benchmark-Pronto-2022language}, FOLIO~\citep{Benchmark-FOLIO2022folio}, and ProverQA~\citep{Benchmark-ProverQA-ICLR-2025large}, is provided in Table~\ref{tab1:dataset}.

\subsection{Discovery of LPT}
\label{sec3.2:discovery}

Using LoCM, we evaluate the performance of LLMs, as shown in Figure~\ref{fig3:LPT}, and observe a previously unrecognized phenomenon: rather than degrading smoothly, reasoning accuracy remains relatively stable over certain complexity ranges and then drops abruptly within specific regions as LoCM increases. 
This collapse behavior consistently appears across both open- and closed-source LLMs. More analysis is provided in Section~\ref{Analysis}.

Analogous to phase transitions in physics~\cite{phase-transitions-1937theory}, where a system exhibits macroscopic changes once a control variable enters a critical region, we term this phenomenon a \textbf{Logical Phase Transition (LPT)} and formalize it as a collapse in model performance governed by logical complexity. 
Importantly, the transition is not characterized by a single threshold, but occurs over one or more \emph{critical intervals}
\[
\mathcal{I}_{k}=[\tau^{(k)}_{\min},\,\tau^{(k)}_{\max}],
\]
within which accuracy drops sharply as $\mathrm{LoCM}(\phi)$ enters the interval and stabilizes again once $\mathrm{LoCM}(\phi)$ exceeds its upper bound. 

The discovery of LPTs indicates that direct exposure to high-complexity samples is ineffective, motivating curriculum learning~\cite{curriculum-learning-2021survey} that organizes samples from easier to harder ones.
By progressively increasing logical complexity, curriculum learning enables stable traversal of transition regions beyond the LPT regimes.

\noindent\textbf{Experience Pool.}
Using the critical intervals from the phase-transition analysis, the dataset is stratified into three regimes:  
Easy samples satisfy $\mathrm{LoCM}(\phi)<\tau^{(1)}_{\min}$;  
Hard samples satisfy $\mathrm{LoCM}(\phi)>\tau^{(K)}_{\min}$, where $K$ is the number of transition intervals;  
all remaining samples fall into the Medium regime.  
This structured \emph{Experience Pool} provides the basis for complexity-aware curriculum scheduling~\cite{curriculum-learning-2021survey} in the Neuro-Symbolic curriculum tuning.

\subsection{Neuro-Symbolic Curriculum Tuning}
\label{sec3.3:tuning}
Motivated by the discovered LPTs, we design a Neuro-Symbolic Curriculum Tuning approach with two complementary components:
(1) \textbf{\textit{Adaptive Neuro-Symbolic Alignment}} for learning a hybrid-semantics model that aligns language and logic representations, and
(2) \textbf{\textit{Complexity-Aware Curriculum}}, which schedules training samples from low to high logical complexity.

\subsubsection*{Adaptive Neuro-Symbolic Alignment}
Hybrid reasoning in LogicAgent~\cite{LogicAgent-2025-ours} shows that NL and FOL provide complementary strengths: NL contributes semantic grounding while FOL supplies precise symbolic constraints. 
Their integration consistently yields higher reasoning accuracy than either modality alone, motivating our hybrid-pretraining strategy.

To obtain a hybrid-semantics model, 
we first fine-tune two baseline models independently: 
a pure NL model \textcolor[HTML]{C04F15}{\(\theta_{\mathrm{NL}}\)} 
and a pure FOL model \textcolor[HTML]{3B7D23}{\(\theta_{\mathrm{FOL}}\)}. 
We then construct a family of hybrid models through linear interpolation:
\begin{equation}
\theta_{\lambda} = (1-\lambda)\,\textcolor[HTML]{C04F15}{\theta_{\mathrm{NL}}}
+ \lambda\,\textcolor[HTML]{3B7D23}{\theta_{\mathrm{FOL}}}, \lambda \in [0,1]
\end{equation}

For each interpolated model \(\theta_{\lambda}\), we fine-tune it on the Neuro-Symbolic alignment dataset using the standard supervised objective. 
We sweep \(\lambda\) on the validation set and select the best-performing configuration as the mixed-semantics model \textcolor[HTML]{1E77B4}{\(\theta_{\mathrm{MIX}}\)}.

\subsubsection*{Complexity-Aware Curriculum Optimization} 

Based on the mixed-semantics model \textcolor[HTML]{1E77B4}{\(\theta_{\mathrm{MIX}}\)}, 
we perform complexity-aware curriculum optimization in which sampling and scheduling are dynamically guided by the observed logical phase-transition behavior. 
Rather than following a fixed Easy$\rightarrow$Medium$\rightarrow$Hard order, 
the model continually monitors its performance relative to the identified critical intervals \(\mathcal{I}_{k}\) detected by the phase-transition analysis.

At each training stage, the model is trained on samples whose complexity $\mathrm{LoCM}(\phi)$ falls within the current target region. 
After each update, performance is re-evaluated to track accuracy changes under the current complexity setting. 
Based on these observations, the curriculum is adaptively adjusted, either continuing optimization within the same region or progressing to higher-complexity samples once gains stabilize. 
This feedback loop ensures that the scheduling is \emph{aware} of the model’s phase-transition behavior during learning.

The model is trained using the standard token-level cross-entropy objective:
\begin{equation}\small
\mathcal{L}(\theta)
= -\, \mathbb{E}_{(x, y_{1:T}) \sim \mathcal{D}}
\Bigg[
\sum_{t=1}^{T}
\log p_{\theta}\!\left(y_t \mid x, y_{<t}\right)
\Bigg]
\end{equation}
where $x$ denotes the input prompt (NL, FOL, or hybrid) and $y_{1:T}$ is the target sequence containing both the reasoning trace and the final answer.

Through this iterative monitoring-and-scheduling process, the model progressively 
stabilizes its reasoning across increasingly difficult regimes, ultimately producing 
the final robust model \textcolor[HTML]{FF800E}{\(\theta^{*}\)}. The complete training procedure is summarized in Algorithm~\ref{alg:phaseaware}.

\algtext*{EndIf}  
\begin{algorithm}[t]
\caption{\textbf{Logical Robustness Training}}
\label{alg:phaseaware}
\small
\begin{algorithmic}[1]

\Require Neuro-Symbolic alignment dataset $\mathcal{D}$

\ForAll{instances $\phi \in \mathcal{D}$}
    \State $\mathrm{LoCM}(\phi) \!\leftarrow\! 
    f\!\left( \sum_{o\in\mathcal{O}}\!\omega(o)\,\mathrm{freq}(o,\phi)+\gamma h(\phi) \right)$ 
\EndFor

\State Detect critical intervals $\{\mathcal{I}_k\}_{k=1}^{K}$
\State Stratify into $\mathcal{D}_{\text{Easy}}, \mathcal{D}_{\text{Med}}, \mathcal{D}_{\text{Hard}}$

\vspace{4pt}
\State \textcolor{CMTgreen}{// Stage 1: Adaptive Neuro-Symbolic Alignment}
\State Train NL model $\textcolor{NLcol}{\theta_{\mathrm{NL}}}$ and FOL model $\textcolor{FOLcol}{\theta_{\mathrm{FOL}}}$ on $\mathcal{D}$

\For{$\lambda \in [0,1]$}
    \State $\theta_{\lambda} \leftarrow (1-\lambda)\,\textcolor{NLcol}{\theta_{\mathrm{NL}}}
    + \lambda\,\textcolor{FOLcol}{\theta_{\mathrm{FOL}}}$
    \State Fine-tune $\theta_{\lambda}$ on $\mathcal{D}$
\EndFor
\State Select best-performing model $\textcolor{MIXcol}{\theta_{\mathrm{MIX}}}$

\vspace{4pt}
\State \textcolor{CMTgreen}{// Stage 2: Complexity-Aware Curriculum Optimization}
\State Initial: $\theta \leftarrow \textcolor{MIXcol}{\theta_{\mathrm{MIX}}}$
\State Evaluate initial overall accuracy $A_{\text{old}} \leftarrow \textsc{Eval}(\theta)$

\For{$i = 1$ to $3$}  \Comment{Easy, Medium, Hard}
    \Repeat
        \State Sample batch $B$ from $\bigcup_{j=1}^{i} \mathcal{D}_j$
        \State Compute loss
        \State $\displaystyle 
        \mathcal{L}(\theta)=
        -\frac{1}{|B|}\!\sum_{(x,y)\in B}
        \sum_{t}\log p_{\theta}(y_t\mid x,y_{<t})$
        \State Update model
        $\theta \leftarrow \theta - \eta\nabla_{\theta}\mathcal{L}(\theta)$
        \State Evaluate overall accuracy $A_{\text{new}}$
        \State Compute $\Delta = A_{\text{new}} - A_{\text{old}}$
        \If{$\Delta > 0$}
            \State Update accuracy $A_{\text{old}} \leftarrow A_{\text{new}}$
        \EndIf
    \Until{$\Delta \le \epsilon$}
\EndFor

\vspace{4pt}
\State \textbf{Output:} final robust model $\textcolor{Finalcol}{\theta^{*}}$

\end{algorithmic}
\end{algorithm}

\section{Experiment}
We organize the experiments into three parts:
 \S~\ref{sec:exp_set} describes the experimental setting;
S~\ref{sec:exp_finetuning} reports the evaluation results;
\S~\ref{sec:exp_ablation} presents ablation studies.

{
\definecolor{lightgreen}{RGB}{220,255,220}
\definecolor{lightred}{RGB}{255,220,220}

\newcommand{\better}[2]{\fcolorbox{white}{lightgreen}{#1}\,{\color{green!60!black}(+#2)}}
\newcommand{\worse}[2]{\fcolorbox{white}{lightred}{#1}\,{\color{red!70!black}(-#2)}}
\newcommand{\same}[1]{#1}

\begin{table*}[h!]\scriptsize
\centering
\caption{
Performance comparison relative to Original. 
\fcolorbox{white}{lightgreen}{Green} = improvement; 
\fcolorbox{white}{lightred}{Red} = degradation; 
\textbf{Bold} = best.
}
\label{tab4:compare_with_baselines}
\renewcommand{\arraystretch}{1.1}
\setlength{\tabcolsep}{6.5pt}

\begin{tabular}{c|l|c|cccccc}
\toprule
\textbf{Method} &
\textbf{Dataset} &
\textbf{Original} &
\textbf{ProntoQA-tuned} &
\textbf{ProofWriter-tuned} &
\textbf{FOLIO-tuned} &
\textbf{ProverQA-tuned} &
\textbf{$\textcolor{Finalcol}{\theta^{*}}$ (Ours)}\\
\midrule
\multirow{6}{*}{\textbf{Naive}}  

& ProntoQA        
  & 55.20 
  & \better{55.40}{0.20}
  & \better{55.40}{0.20}
  & \better{56.20}{1.00}
  & \worse{54.80}{0.40}
  & \textbf{\better{56.80}{1.60}}
\\

& ProofWriter     
  & 44.16 
  & \worse{44.00}{0.16}
  & \better{44.33}{0.17}
  & \worse{43.80}{0.36}
  & \better{44.30}{0.14}
  & \textbf{\better{44.66}{0.50}}
\\

& FOLIO           
  & 60.78
  & \worse{59.80}{0.98}
  & \worse{59.80}{0.98}
  & \better{61.27}{0.49}
  & \worse{59.80}{0.98}
  & \textbf{\better{62.25}{1.47}}
\\

& ProverQA        
  & 54.13
  & \worse{54.07}{0.06}
  & \worse{53.20}{0.93}
  & \worse{53.80}{0.33}
  & \better{54.73}{0.60}
  & \textbf{\better{55.47}{1.34}}
\\

& NSA-LR     
  & 49.55
  & \worse{49.09}{0.46}
  & \worse{49.54}{0.01}
  & \worse{49.09}{0.46}
  & \better{50.45}{0.90}
  & \textbf{\better{50.91}{1.36}}
\\
\cmidrule(lr){2-8}
& \textbf{Naive Ave} 
  & \same{52.76}
  & \worse{52.47}{0.29}
  & \worse{52.45}{0.31}
  & \better{52.83}{0.07}
  & \better{52.82}{0.06}
  & \textbf{\better{54.02}{1.26}}
\\

\midrule
\multirow{6}{*}{\textbf{CoT}}  

& ProntoQA
  & 67.60
  & \better{69.40}{1.80}
  & \better{70.60}{3.00}
  & \better{71.40}{3.80}
  & \better{71.20}{3.60}
  & \textbf{\better{72.00}{4.40}}
\\

& ProofWriter
  & 55.16
  & \worse{54.33}{0.83}
  & \worse{54.00}{1.16}
  & \better{56.33}{1.17}
  & \worse{53.50}{1.66}
  & \textbf{\better{60.71}{5.55}}
\\

& FOLIO
  & 66.17
  & \worse{65.68}{0.49}
  & \worse{63.23}{2.94}
  & \worse{63.72}{2.45}
  & \textbf{\better{66.67}{0.50}}
  & \worse{65.20}{0.97}
\\

& ProverQA 
  & 60.70
  & \worse{55.70}{5.00}
  & \better{60.80}{0.10}
  & \worse{60.70}{0.00}
  & \better{61.50}{0.80}
  & \textbf{\better{64.20}{3.50}}
\\

& NSA-LR
  & 57.70
  & \better{58.20}{0.50}
  & \worse{54.70}{3.00}
  & \better{58.63}{0.93}
  & \better{59.62}{1.92}
  & \textbf{\better{65.00}{7.30}}
\\
\cmidrule(lr){2-8}
& \textbf{CoT Ave} 
  & \same{61.47}
  & \worse{60.66}{0.81}
  & \worse{60.67}{0.80}
  & \better{62.16}{0.69}
  & \better{62.50}{1.03}
  & \textbf{\better{65.42}{3.95}}
\\

\bottomrule
\end{tabular}
\end{table*}

\begin{table}[t]
\centering
\small
\setlength{\tabcolsep}{4.5pt}
\caption{Comparison with recent training-free reasoning methods on NSA-LR.}
\label{tab:training_free_comparison}
\begin{tabular}{@{}lcccc@{}}
\toprule
Method & Low & Medium & High & Overall \\
\midrule

Na\"ive
& 60.6 & 49.8 & 38.6 & 49.6 \\
\hspace{0.7em}+ NSCT
& \cellcolor{lightgreen}60.9{\scriptsize\,(+0.3)}
& \cellcolor{lightgreen}50.2{\scriptsize\,(+0.4)}
& \cellcolor{lightgreen}41.6{\scriptsize\,(+3.0)}
& \cellcolor{lightgreen}50.9{\scriptsize\,(+1.3)} \\
\midrule

CoT
& 75.5 & 58.4 & 39.4 & 57.7 \\
\hspace{0.7em}+ NSCT
& \cellcolor{lightgreen}84.0{\scriptsize\,(+8.5)}
& \cellcolor{lightgreen}64.2{\scriptsize\,(+5.8)}
& \cellcolor{lightgreen}46.8{\scriptsize\,(+7.4)}
& \cellcolor{lightgreen}65.0{\scriptsize\,(+7.3)} \\
\midrule

ToT
& 82.0 & 64.8 & 42.4 & 63.2 \\
\hspace{0.7em}+ NSCT
& \cellcolor{lightgreen}88.6{\scriptsize\,(+6.6)}
& \cellcolor{lightgreen}68.3{\scriptsize\,(+3.5)}
& \cellcolor{lightgreen}48.5{\scriptsize\,(+6.1)}
& \cellcolor{lightgreen}68.4{\scriptsize\,(+5.2)} \\
\midrule

DetermLR
& 84.0 & 63.9 & 43.8 & 63.9 \\
\hspace{0.7em}+ NSCT
& \cellcolor{lightgreen}87.5{\scriptsize\,(+3.5)}
& \cellcolor{lightgreen}67.9{\scriptsize\,(+4.0)}
& \cellcolor{lightgreen}49.2{\scriptsize\,(+5.4)}
& \cellcolor{lightgreen}68.2{\scriptsize\,(+4.3)} \\
\midrule

SymbCoT
& 83.8 & 63.7 & 46.9 & 64.8 \\
\hspace{0.7em}+ NSCT
& \cellcolor{lightgreen}89.2{\scriptsize\,(+5.4)}
& \cellcolor{lightgreen}68.5{\scriptsize\,(+4.8)}
& \cellcolor{lightgreen}50.5{\scriptsize\,(+3.6)}
& \cellcolor{lightgreen}69.5{\scriptsize\,(+4.7)} \\
\bottomrule
\end{tabular}
\end{table}
}

\subsection{Settings}
\label{sec:exp_set}

\noindent\textbf{Model.} For \emph{tuning}, we adopt Qwen2.5-7B as the base model and apply parameter-efficient LoRA tuning. The model is trained with HuggingFace TRL on a single GPU using fp16 quantization. We use a learning rate of $1\times10^{-4}$, one training epoch, a per-device batch size of 1, and \texttt{gradient\_accumulation\_steps} = 48. For \emph{LPTs analysis}, we assess LLMs including the open-source Qwen2.5 (3B/7B/14B/32B), Qwen3 (1.7B/4B/8B/14B/30B), and Gemma (1B/4B/12B/27B) families, as well as the closed-source GPT-4.1 Nano/Mini and DeepSeek V3.1. All evaluations use temperature=0 decoding.

\noindent\textbf{Symbolic Toolkit.} To verify the syntactic validity of FOL forms, we employ the \texttt{nltk}~\citep{tool-nltk-2006nltk} library for CFG-based structural checking.

\noindent\textbf{Benchmarks.} 
We evaluate on four established logical reasoning benchmarks: 
ProntoQA~\citep{Benchmark-Pronto-2022language} (5-hop subset), 
ProofWriter~\citep{Benchmark-ProofWriter-2020proofwriter} (depth-5, OWA setting), 
FOLIO~\citep{Benchmark-FOLIO2022folio} (full expert-curated split), and 
ProverQA~\citep{Benchmark-ProverQA-ICLR-2025large} (1500 examples in  total, 1--9 reasoning steps). 
Detailed benchmark descriptions are provided in Appendix~\ref{app:dataset_details}.

\noindent\textbf{Baselines.}
We evaluate six model variants built upon Qwen2.5--7B: 
(1) the original model (\emph{Orig.}), 
(2) a model fine-tuned on ProntoQA data, 
(3) a model fine-tuned on ProofWriter data, 
(4) a model fine-tuned on FOLIO data, 
(5) a model fine-tuned on ProverQA data, 
and (6) our multi-stage fine-tuned model ($\textcolor{Finalcol}{\theta^{*}}$). 
For each model variant, we compare two prompting baselines: 
\emph{Naive Prompting} and 
\emph{Chain-of-Thought (CoT)}~\citep{COT-2022chain}. 

\begin{figure*}[t]
\centering
\includegraphics[width=\linewidth]{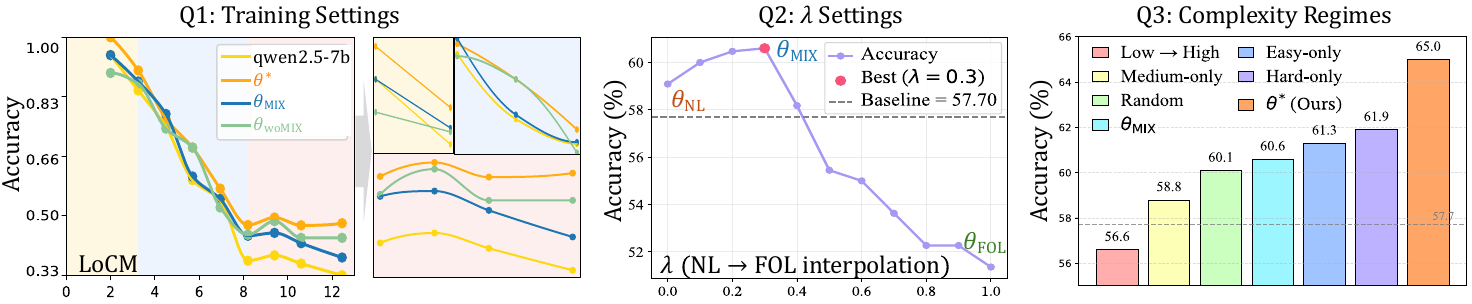}
\caption{
Ablation studies of the proposed multi-stage training framework.
\textbf{(Left: Q1)} Comparison of curriculum learning and representation mixing.
\textbf{(Middle: Q2)} Sensitivity analysis with respect to the mixing coefficient $\lambda$.
\textbf{(Right: Q3)} Performance under different logical complexity regimes.
}
\label{fig:ablation_all}
\end{figure*}

\subsection{Evaluation Results}
\label{sec:exp_finetuning}

Tables~\ref{tab4:compare_with_baselines} and \ref{tab:training_free_comparison} report the evaluation results, from which we draw three main observations.

\noindent\textbf{(1) Our method outperforms alternative fine-tuning strategies.}
Under Naive prompting, $\textcolor{Finalcol}{\theta^{*}}$ is the only variant that improves over the Original model, achieving the best score on every evaluated dataset with an average gain of (+1.26).
In contrast, single-dataset fine-tuning yields only marginal improvements or even degradation, indicating that LoCM-guided multi-stage training is more effective at mitigating reasoning collapse.
Under CoT prompting, despite increased reasoning difficulty, $\textcolor{Finalcol}{\theta^{*}}$ again achieves the highest average performance (+3.95) among all variants.
While other methods exhibit mixed gains or notable drops on specific datasets, $\textcolor{Finalcol}{\theta^{*}}$ consistently attains the best or near-best results.

\noindent\textbf{(2) Our method on NSA-LR generalizes to heterogeneous reasoning datasets.}
Our model $\textcolor{Finalcol}{\theta^{*}}$, trained via Neuro-Symbolic Curriculum Tuning on the NSA-LR dataset, achieves strong performance on ProntoQA, ProofWriter, FOLIO, and ProverQA, consistently ranking among the top results.
In contrast, alternative fine-tuning methods that are optimized on individual datasets exhibit limited gains on their training datasets and often suffer noticeable performance degradation when evaluated on other reasoning datasets, failing to generalize across heterogeneous tasks.
These results indicate that curriculum tuning guided by neuro-symbolic alignment captures transferable reasoning structure, enabling performance improvements that extend beyond the training dataset to diverse logical forms and complexity regimes. One exception is FOLIO under CoT prompting, where the improvement is not maintained; we provide further analysis in Appendix~C.x.

\noindent\textbf{(3) Our method consistently enhances diverse inference-time reasoning strategies.}
As shown in Table~\ref{tab:training_free_comparison}, NSCT improves performance not only under Naive and CoT prompting, but also when combined with recent training-free reasoning methods such as ToT, DetermLR, and SymbCoT. 
These gains are consistent across all settings, indicating that the effect of NSCT does not depend on a specific prompting format or decoding heuristic. 
Notably, the improvements are most pronounced in the High-LoCM regime, where logical phase transitions are most severe: under CoT, ToT, and DetermLR, the gains in the High-LoCM band are +7.4, +6.1, and +5.4, respectively, all larger than those in the Low- or Medium-LoCM bands. 
Overall, these results support the view that NSCT is not merely a narrow fine-tuning heuristic tied to a specific prompting strategy. Instead, it behaves as a transferable training mechanism that consistently improves logical robustness across diverse decoding settings.

\subsection{Ablation Study}
\label{sec:exp_ablation}
To evaluate our method, we conduct ablation experiments to answer the following questions:
\begin{enumerate}[left=0pt, topsep=2pt, itemsep=0pt]
  \item \textbf{Is combining neuro-symbolic alignment with curriculum optimization effective?}
  \item \textbf{How sensitive is the method to the interpolation coefficient $\lambda$?}  
  \item \textbf{How does training on different complexity regimes affect performance?}  
\end{enumerate}

\noindent\textbf{Q1: Impact of combining neuro-symbolic alignment and curriculum optimization.}  
To address \textbf{Q1}, we compare four variants (Figure~\ref{fig:ablation_all}-Q1):  
(1) the base model,  
(2) \textcolor[HTML]{97C397}{\(\theta_{\mathrm{woMIX}}\)}, which applies curriculum optimization without neuro-symbolic alignment,  
(3) \textcolor[HTML]{1E77B4}{\(\theta_{\mathrm{MIX}}\)}, which uses neuro-symbolic alignment without curriculum optimization,  
and (4) our full model \textcolor{Finalcol}{\(\theta^{*}\)}, which integrates both components.

\textbf{(1) All variants exhibit consistent LPT trends.}
Across all variants, reasoning accuracy follows a similar trend as LoCM increases, with a sharp collapse beyond critical thresholds and subsequent convergence to random guessing.
This indicates that LPTs are intrinsic to the underlying model and are not eliminated by neuro-symbolic alignment or curriculum optimization.

\textbf{(2) Fine-tuning improves accuracy at matched complexity levels, with the combined model performing best.}
Although the overall LPT pattern remains unchanged, all variants outperform the base model at comparable complexity levels.
The full model \textcolor{Finalcol}{\(\theta^{*}\)} consistently exceeds
\textcolor[HTML]{1E77B4}{\(\theta_{\mathrm{MIX}}\)} and \textcolor[HTML]{97C397}{\(\theta_{\mathrm{woMIX}}\)},
indicating that neuro-symbolic alignment and curriculum optimization provide complementary benefits within each complexity regime.

\paragraph{Q2: Impact of balancing FOL and NL supervision (\(\lambda\)).}  
We sweep \(\lambda\) between pure NL supervision (\(\lambda=0\)) and pure FOL supervision (\(\lambda=1\)) to examine how the two modalities contribute to reasoning behavior (Figure~\ref{fig:ablation_all}-Q2).  

\textbf{(1) Pure NL improves over baseline, while pure FOL severely underperforms.} 
At \(\lambda=0\), training with natural language alone already surpasses the untuned baseline, indicating that NL supervision enhances reasoning by preserving contextual and discourse-level cues.  
In contrast, at \(\lambda=1\), pure FOL training causes a marked accuracy drop: while enforcing strict symbolic structure, it removes lexical semantics, pragmatic signals, and implicit assumptions required to interpret natural-language queries.  
As a result, the model aligns tightly with formal expressions but fails to robustly map them back to their natural-language realizations, leading to degraded semantic reasoning under FOL-only supervision.

\textbf{(2) Moderate NL–FOL mixing yields the best trade-off.}  
Accuracy peaks around \(\lambda=0.3\), where natural language continues to provide semantic grounding and flexible understanding of varied phrasings, while FOL supplies explicit operator-level structure and removes many logically inconsistent patterns.  
In this regime, symbolic precision from FOL and semantic richness from NL are complementary rather than competing, leading to the strongest overall performance and more robust generalization across logically complex queries.

\begin{table}[t]\small
\centering
\caption{Accuracy–complexity correlation under different monotone transforms.}
\label{tab:transform-corr}
\begin{tabular}{l l c}
\toprule
\textbf{Transform} & \textbf{Definition} & \textbf{Corr.} \\
\midrule
Linear   & \(f(C) = C\)          & $-0.391$ \\
Log      & \(f(C) = \log(C+1)\)  & $-0.384$ \\
Square   & \(f(C) = C^2\)        & $-0.343$ \\
Inverse  & \(f(C) = 1/(C+1)\)    & $+0.248$ \\
\textbf{Sqrt} & \(\boldsymbol{f(C) = \sqrt{C}}\) & \(\mathbf{-0.399}\) \\
\bottomrule
\end{tabular}
\end{table}


\paragraph{Q3: Effect of training under different complexity regimes.}
We examine how different complexity regimes and training schedules influence model performance (Figure~\ref{fig:ablation_all}-Q3).

\textbf{(1) Single-regime fine-tuning yields limited gains.}
Training on a single complexity regime yields only marginal improvements over the mixed baseline ($60.6$), with accuracies of $61.3$ (easy-only), $58.8$ (medium-only), and $61.9$ (hard-only), indicating that no individual regime alone supports robust generalization.
Notably, medium-only training performs the worst ($58.8$), reflecting less stable learning signals.

\textbf{(2) Naive low-to-high training causes forgetting, while complexity-aware curriculum is effective.}
A low-to-high training strategy leads to a substantial performance drop ($56.6$) due to catastrophic forgetting, as earlier learned patterns are overwritten by harder instances.
In contrast, our complexity-aware curriculum achieves the best performance ($65.0$) by retaining earlier regimes while progressively introducing higher complexity, thereby enabling consistent gains across complexity levels. More details in Appendix~\ref{app:curriculum_necessity}.

\subsection{Analysis and Discussion}
\label{Analysis}
We discuss the following three aspects:
\begin{enumerate}[left=0pt, topsep=2pt, itemsep=0pt]
    \item \textbf{Decomposition of LoCM.}  
    We analyze the operator weights and nonlinearity \(f(\cdot)\) in LoCM.
    
    \item \textbf{LPTs in LLMs.}  
    We examine the emergence of LPTs in LLMs across LoCM ranges.
    
    \item \textbf{Structured prompting on LPTs.}  
    We discuss how CoT prompting influences LPTs.
\end{enumerate}


\paragraph{D1: Decomposition of LoCM.} Based on the empirical results, we make the following observations. Additional analyses are provided in Appendix~\ref{sec:appendix_metric_diag}.

\noindent\textbf{(1) Operator-weight justification.}  
As shown in Table~\ref{tab:LoCM_weight_corr}, assigning weights to logical operators is necessary to reflect their symbolic roles. Removing or isolating any operator type consistently weakens the correlation between complexity and accuracy, indicating that no single connective dominates reasoning difficulty. The weighted formulation is therefore required to capture interaction structures among logical operators.

\noindent\textbf{(2) Choice of nonlinearity \(f(\cdot)\).}  
As shown in Table~\ref{tab:transform-corr}, we compare monotone transformations applied to aggregated operator contributions. The square-root mapping achieves the strongest alignment with accuracy, while logarithmic and square transformations introduce distortions in low- or high-complexity regions. These trends motivate selecting the square-root function as the final instantiation of \(f(\cdot)\).

\begin{table}[t]
\centering
\small
\setlength{\tabcolsep}{6pt}
\renewcommand{\arraystretch}{1.05}
\caption{
Accuracy–complexity correlation under LoCM operator ablations.
}
\label{tab:LoCM_weight_corr}
\begin{tabular}{lccc}
\toprule
\textbf{Setting} & \textbf{Operator(s)} & \textbf{Type} & \textbf{Corr.} \\ 
\midrule

\multirow{6}{*}{Remove}
& $\lnot$ & Negation & -0.377 \\
& $\land,\lor$ & Basic connectives & -0.366 \\
& $\forall,\exists$ & Quantifiers & -0.374 \\
& $\rightarrow,\leftrightarrow$ & Conditional & -0.370 \\
& $\oplus$ & XOR & -0.368 \\
& $h$ & Hops & -0.385 \\

\midrule

\multirow{6}{*}{Only}
& $\lnot$ & Negation & -0.263 \\
& $\land,\lor$ & Basic connectives & -0.358 \\
& $\forall,\exists$ & Quantifiers & -0.289 \\
& $\rightarrow,\leftrightarrow$ & Conditional & -0.352 \\
& $\oplus$ & XOR & -0.325 \\
& $h$ & Hops & -0.370 \\
\midrule
Full & All operators & Complete LoCM & \textbf{-0.391} \\
\bottomrule
\end{tabular}
\end{table}

\paragraph{D2: LPTs in LLMs.}
Figure~\ref{fig3:LPT} shows several consistent phenomena across open- and closed-source LLMs; see Appendix~\ref{app:full_results} for full results.

\noindent\textbf{(1) Logical phase transitions universally occur across models.}
As LoCM increases, all evaluated models exhibit sharp accuracy drops, typically forming one or two distinct transition intervals. This demonstrates that phase-transition behavior is a fundamental property of current LLMs rather than an artifact of model family or training paradigm.

\noindent\textbf{(2) Accuracy eventually converges to the random-guessing baseline.}
Once the instance LoCM surpasses the upper bound of the final transition interval, model accuracy approaches the $1/3$ level indicated by the dashed line, implying that final predictions are essentially guesses rather than results of successful intermediate reasoning.

\noindent\textbf{(3) Larger models exhibit delayed degradation and stronger robustness.}
In the Gemma~3 family (green curves), larger models maintain higher accuracy across the full complexity range, reflecting uniformly improved robustness rather than a mere horizontal shift of the transition region. In contrast, the 1B model collapses once LoCM exceeds~8 and quickly approaches the random baseline, showing its inability to handle high-complexity reasoning.

\begin{figure}[t]
    \centering
    \includegraphics[width=1.0\linewidth]{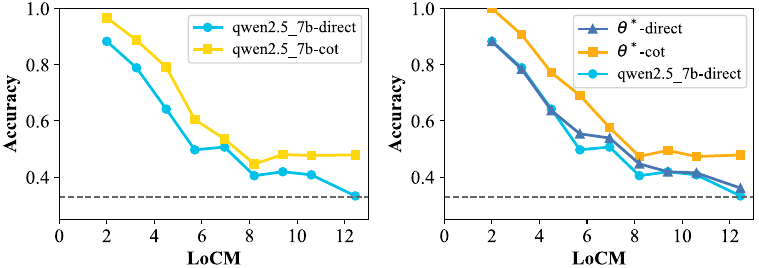}
    \caption{Effect of prompting strategies on logical phase transition (LPT) curves.
    \textbf{Left:} Base model (Qwen2.5-7B, without fine-tuning) under Direct and CoT prompting.
    \textbf{Right:} Comparison between the base model and the finetuned model under Direct and CoT prompting.}
    \label{fig:compare_cot_tuning}
\end{figure}

\paragraph{D3: Structured prompting on LPTs.}
As shown in Figure~\ref{fig:compare_cot_tuning}, both fine-tuning and CoT prompting consistently raise accuracy across logical-complexity bins.
In particular, CoT yields uniform gains over Direct prompting at nearly all complexity levels, indicating improved inference efficiency and capacity utilization.
However, despite these gains, the onset of the logical phase transition remains unchanged, and performance still degrades sharply once complexity exceeds the model’s intrinsic limit.
Therefore, \textbf{neither fine-tuning nor structured prompting extends the complexity horizon; instead, they improve performance strictly within the same phase regime}.

\section{Conclusion}
We introduce the Logical Complexity Metric (LoCM) as a principled, model-agnostic measure for characterizing the limits of logical reasoning in large language models, and show that LLMs exhibit universal logical phase transitions marked by abrupt accuracy collapse once complexity exceeds a critical threshold. Guided by this diagnostic insight, we develop a complexity-graded reasoning corpus and a neuro-symbolic curriculum optimization that aligns training with complexity progression. Experiments demonstrate that this approach consistently improves robustness across diverse benchmarks and prompting settings, mitigating post-transition failure and strengthening high-complexity logical reasoning.

\section*{Limitations}
\textbf{(1) Logical phase transitions persist under fine-tuning and structured prompting.}
Neither fine-tuning nor CoT prompting delays, shifts, or eliminates LPTs.
Although both improve accuracy within fixed complexity regimes, the critical thresholds remain unchanged, and performance still collapses sharply beyond them.
This indicates that LPTs arise from intrinsic properties of current model architectures and inference mechanisms rather than from deficiencies addressable by tuning or prompting.
\textbf{(2) Dependence on existing datasets and a focus on symbolic-dominant reasoning.}
Our framework relies on re-annotating existing logical reasoning datasets with complete FOL representations, rather than introducing a new data generation paradigm, which constrains the diversity of logical forms to the underlying corpora.
Moreover, LoCM targets reasoning expressible in FOL and does not directly cover reasoning driven by commonsense knowledge, world modeling, or probabilistic inference.
(3)\textbf{Phase-transition boundaries require model-specific empirical calibration.}
Although LPT phenomena are consistently observed across both open- and closed-source LLMs, the exact locations and shapes of critical intervals vary across model families and must be empirically calibrated.
This suggests that LoCM captures relative rather than absolute difficulty, and that transferring phase boundaries across architectures should be done with caution.

\section*{Acknowledgments}
This work is supported by the National Natural Science Foundation of China (Numbers 62272184 and 62402189), 
the China Postdoctoral Science Foundation (Numbers 2024M751012, 2025T180429, and GZC20230894),  
the Postdoctor Project of Hubei Province (Number 2024HBBHCXB014),
the Natural Science Foundation of Hubei Province No.JCZRMS202600758,
and Sponsored by CIPS-SMP-Zhipu Large Model Fund (CIPS-SMP20250306),
The computation is completed in the HPC Platform of Huazhong University of Science and Technology.

\bibliography{main}

\appendix
\clearpage
\section*{\centering Appendix}

\noindent
The appendix is organized as follows.
\S~\ref{app:related_work} reviews additional related work.
\S~\ref{appendix:datasets_and_baselines} describes datasets and baselines.
\S~\ref{sec:appendix_experimental_details} provides experimental details.
\S~\ref{app:formal_setup} presents the formal setup.
\S~\ref{app:future_directions} discusses future directions.
\S~\ref{sec:full-prompting} includes full prompting templates.

\section*{The Usage of LLM}
In accordance with ACL guidelines, we used large language models solely for writing assistance and language refinement. 

\section{Related Work}
\label{app:related_work}
\subsection{Benchmarks for Logical Reasoning}

Existing logical reasoning benchmarks fall broadly into two categories. 
\textbf{Synthetic symbolic benchmarks} such as PrOntoQA~\citep{Benchmark-Pronto-2022language} and ProofWriter~\citep{Benchmark-ProofWriter-2020proofwriter} evaluate deduction in controlled symbolic settings using rule-based or template-driven generation. 
\textbf{Natural-language logical benchmarks} such as FOLIO~\citep{Benchmark-FOLIO2022folio} and ProverQA~\citep{Benchmark-ProverQA-ICLR-2025large} provide English scenarios paired with verified reasoning chains produced via theorem-proving pipelines.  
Together, these datasets offer structured environments for assessing the logical validity of model outputs.

\textbf{Open Challenges.}
Despite their utility, existing benchmarks often emphasize \emph{logical form} in relatively controlled settings, typically using concrete and unambiguous predicates with fixed semantics. 
They may not provide complete end-to-end symbolic representations (e.g., full FOL contexts paired with verified reasoning chains) under a unified schema at scale.  
As a result, they offer limited support for examining how reasoning difficulty scales with increasing \emph{logical complexity}, motivating our construction of a complexity-graded Neuro-Symbolic alignment dataset.

\subsection{LLM-Based Logical Reasoning}

\textbf{Reasoning Methodologies.}
Prior work on LLM~\cite{liu2025fusion, hu2025sf2t,ye2025mvp,zeng2026vision,han2026unicorn, song11} logical reasoning can be grouped into three main directions:
(1) \textbf{Linear Reasoning (LR)} methods such as Naive Prompting and Chain-of-Thought~\citep{COT-2022chain,chen-etal-2025-improving-reasoning, lin2026mmfinereason}, which generate step-by-step natural-language explanations;
(2) \textbf{Aggregative Reasoning (AR)} approaches~\citep{CLOVER-ICLR-2024divide, Cumulative-reasoning-2023-TsinghuIIIS, TOT-2023tree, Determlr-ACL-2024-RenDaGaoling, song13, song4} that combine multiple reasoning trajectories through sampling, search, or majority voting; and
(3) \textbf{Symbolic Reasoning (SR)} frameworks~\citep{NL2FOL-2023harnessing, Baseline-Aristotle2024xu, Symbolic-COT-2024faithful, LogicLM-2023} that integrate LLMs~\citep{ qwen2.5-2025qwen2,wang2026PERM,wang2026reasoning,li2025llm} with explicit logic modules or FOL-based reasoning.
Together, these methodologies improve performance on a wide range of reasoning benchmarks through guided, multi-path, or logic-augmented reasoning.

\textbf{Open Challenges.}
Despite these advances, many methodologies primarily improve prompting, reasoning-time scaffolds, or external reasoning modules~\cite{ling2026neuralchainofthoughtsearchsearching,pcgr2025naacl}, rather than explicitly characterizing or strengthening robustness across varying levels of \emph{logical complexity}. 
As a result, most approaches improve accuracy on fixed benchmarks but provide limited insight into how internal reasoning behavior changes along a fine-grained complexity continuum, which is the central focus of our work.

\begin{table*}[h]
\centering
\small
\caption{Comparison of dataset statistics. NSA-LR scales up the ProverQA framework and adds verifiable symbolic reasoning chains. $^{\dagger}$\textit{Statistics for ProofWriter refer to the depth-5 OWA subset used in our experiments.}}
\label{tab:dataset_comparison}
\resizebox{\textwidth}{!}{%
\newcommand{\cmark}{\textcolor{green!60!black}{\ding{51}}}
\newcommand{\xmark}{\textcolor{red!70!black}{\ding{55}}}
\begin{tabular}{lccccccccc}
\toprule
\textbf{Dataset} & \textbf{Method} & \textbf{Scalability} & \textbf{Rich NL} & \textbf{Sym. Rep.} & \textbf{Verif. Chain} & \textbf{Verif. Chain} & \textbf{NL-Sym} & \textbf{Test Set} & \textbf{Train Set} \\
 & & & & & \textbf{(NL)} & \textbf{(Sym.)} & \textbf{Mapping} & \textbf{Size} & \textbf{Size} \\
\midrule
\textbf{ProverQA} & Synthetic & \cmark & \cmark & \cmark & \cmark & \xmark & \cmark & 1,500 & 5,000 \\
\textbf{FOLIO} & Manual & \xmark & \cmark & \cmark & \xmark & \xmark & \xmark & 204 & 1,000 \\
\textbf{ProofWriter$^{\dagger}$} & Synthetic & \cmark & \xmark & \xmark & \cmark & \xmark & \xmark & 600 & 5,000 \\
\textbf{ProntoQA} & Synthetic & \cmark & \xmark & \cmark & \cmark & \xmark & \xmark & 500 & 2,880 \\
\midrule
\textbf{NSA-LR (Ours)} & Synthetic & \cmark & \cmark & \cmark & \cmark & \cmark & \cmark & 1,500 & \textbf{15,800} \\
\bottomrule
\end{tabular}%
}
\end{table*}

\subsection{Logical Reasoning Capacity Analysis}
\label{A.3}
\textbf{Existing Capacity and Fidelity Analyses.}
A substantial body of work investigates the reasoning~\citep{li2026agent,li2026GRPO,hao2025rethinking,hao2026recreate,stage} fidelity and compositionality of LLMs~\cite{GAS3-2025,li2024coupled,song2024autogenic,chen2026transformer, cineagent, zhou2026look}. 
Prior studies show that models often hallucinate intermediate steps or deviate from valid derivations even when answers are correct~\citep{Reasoning-survey-2022towards, multi-step-2024exploring}, revealing instability in multi-step reasoning and difficulty maintaining consistent logical chains. 
More recent capacity-oriented research examines how reasoning behavior changes as tasks grow more complex. 
InfoQA~\citep{InfoQA-ICLR2025-XiuyingChen-MBZUAI} formalizes a single-pass accuracy upper bound for multi-hop QA, demonstrating that performance can collapse once information load exceeds model capacity. 
CoT-Valve~\citep{CoT-Valve-ACL2025-NUS} shows that reasoning chains can be compressed or expanded through latent control, highlighting sensitivity to chain length and internal reasoning load. 
Symbolic Monte Carlo supervision~\citep{Sym-MC-EMNLP2025-Sheffield} leverages symbolic trajectories to stabilize intermediate reasoning.  

\textit{Apple}~\citep{illusion-of-thinking-Nips2025-Apple} further demonstrates abrupt accuracy collapses in structured puzzles such as Tower of Hanoi as complexity increases.

\textbf{Open Challenges.}
Although informative, existing analyses often rely on coarse difficulty indicators such as hop count, chain length, or rule depth, offering only partial insight into how reasoning performance evolves under more nuanced forms of complexity. 
They typically lack a unified framework for quantifying \emph{logical complexity} and for characterizing how LLM behavior changes across fine-grained complexity levels. 
Our work complements these analyses by introducing an LoCM and systematically evaluating model performance along a controlled complexity continuum.

\section{Dataset Details}
\label{appendix:datasets_and_baselines}
\subsection{Dataset Descriptions}
\label{app:dataset_details}

We describe the logical reasoning datasets used in our evaluation, which together provide a systematic and fair testbed for assessing model reasoning across diverse logical structures and deductive complexity levels.

\paragraph{FOLIO} FOLIO is an expert-curated benchmark for natural language reasoning with first-order logic (FOL) annotations. Compared with synthetic datasets, it offers greater linguistic diversity and naturalness while preserving strict logical validity. The task requires determining whether a conclusion follows from a set of premises, involving phenomena such as quantification and negation, and thus serves as a key testbed for evaluating the integration of semantic understanding and formal reasoning.

\paragraph{ProofWriter} ProofWriter is a synthetic benchmark designed to evaluate systematic neural deduction under the Open World Assumption. It features rule-based natural language problems requiring multi-step reasoning over conjunctions and disjunctions. Its explicit proof supervision enables evaluation of both final-answer correctness and the validity of intermediate deductive steps.

\paragraph{ProntoQA} ProntoQA is a synthetic benchmark for diagnosing deductive reasoning in a highly controlled symbolic setting. Each instance contains premises, a step-by-step chain of thought, and a conclusion, allowing precise analysis of multi-hop inference while minimizing linguistic ambiguity.

\paragraph{ProverQA} ProverQA, built on ProverGen, combines LLM-generated linguistic diversity with formally verified theorem-proving chains. It is designed to assess deductive consistency and symbolic–linguistic alignment, with stratified reasoning depths that support scalable evaluation across increasing inference complexity.

\begin{figure*}[t]
    \centering
    \includegraphics[width=\linewidth]{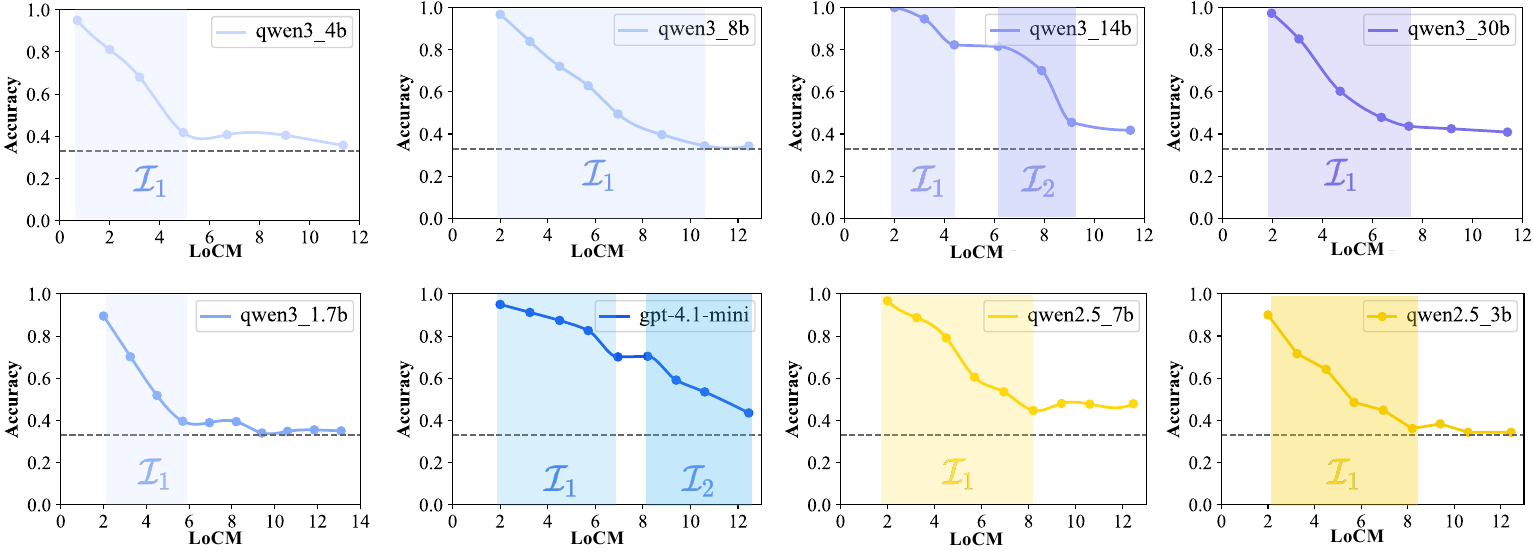}
    \caption{
    Supplementary logical phase-transition curves for additional models.
    Shaded bands indicate the detected transition intervals (e.g., $\mathcal{I}_1$, $\mathcal{I}_2$), where accuracy exhibits an abrupt decline as LoCM increases.
    The dashed horizontal line denotes the $1/3$ random-guess baseline.
    }
    \label{fig:Supplementary_LPT}
\end{figure*}

\paragraph{NSA-LR (Ours).}
To enable fine-grained logical complexity analysis, we construct a Neuro-Symbolic alignment dataset extending the ProverGen framework used in ProverQA.
Unlike the original framework, which focuses on verifying reasoning chains, our key modification exhaustively translates all natural language components, including propositions, premise sets, and intermediate reasoning steps, into explicit first-order logic.
By strictly aligning each sentence with its logical form using unified translation rules, the dataset provides the symbolic grounding required to compute LoCM and systematically analyze how structural logical complexity affects model performance.

\subsection{Dataset Statistics and Comparison}
\label{app:dataset_stats}

Table~\ref{tab:dataset_comparison} presents the statistics and key characteristics of the datasets. While FOLIO provides mappings between natural language and FOL, it lacks verifiable reasoning chains. In contrast, our \textbf{NSA-LR} dataset (an extension of ProverQA) uniquely combines scalability, linguistic richness, and fully verifiable symbolic reasoning chains, supporting rigorous logic-consistency analysis.

\subsection{Example of NSA-LR Data Instance}
\label{app:fol_plus_example}

For a detailed walkthrough of a high-complexity reasoning instance ($LoCM \approx 7.25$) that demonstrates the strict alignment between natural language and first-order logic across a multi-step deduction chain, please refer to the case study presented in Figure~\ref{fig:kaizen_case_study}.




\begin{table*}[t]
\centering
\small

\definecolor{myblue1}{rgb}{0.992, 0.966, 0.847}    
\definecolor{myblue2}{rgb}{0.908, 0.915, 0.992}
\definecolor{othercolor}{rgb}{0.822, 0.911, 0.985}
\definecolor{mygreen1}{rgb}{0.900, 0.983, 0.875}

\setlength{\tabcolsep}{12pt}
\renewcommand{\arraystretch}{1.3}

\caption{
Full results (accuracy) across model families and complexity bins. 
Performance systematically degrades from Bin 1 to Bin 9 as task complexity increases.
}
\label{tab:full_results_model_by_bins}

\begin{tabular}{@{}l*{9}{S[table-format=1.3]}@{}}
\toprule
\textbf{Model} & 
\multicolumn{9}{c@{}}{\textbf{Complexity Bins}} \\
\cmidrule(lr){2-10}
& {\textbf{1}} & {\textbf{2}} & {\textbf{3}} & {\textbf{4}} & 
{\textbf{5}} & {\textbf{6}} & {\textbf{7}} & {\textbf{8}} & {\textbf{9}} \\
\midrule

\rowcolor{myblue1} 
\multicolumn{10}{@{}l}{\small\textit{Qwen 2.5 Series}} \\
\addlinespace[2pt]
Qwen2.5-3B    & 0.900 & 0.716 & 0.642 & 0.485 & 0.448 & 0.361 & 0.383 & 0.343 & 0.343 \\
Qwen2.5-7B    & 0.950 & 0.851 & 0.749 & 0.600 & 0.543 & 0.374 & 0.389 & 0.366 & 0.333 \\
Qwen2.5-14B   & 0.967 & 0.943 & 0.894 & 0.846 & 0.689 & 0.537 & 0.504 & 0.512 & 0.362 \\
Qwen2.5-32B   & 0.967 & 0.923 & 0.837 & 0.821 & 0.669 & 0.544 & 0.480 & 0.403 & 0.404 \\
\addlinespace[6pt]

\rowcolor{myblue2} 
\multicolumn{10}{@{}l}{\small\textit{Qwen 3 Series}} \\
\addlinespace[2pt]
Qwen3-1.7B    & 0.895 & 0.702 & 0.518 & 0.396 & 0.389 & 0.394 & 0.340 & 0.348 & 0.353 \\
Qwen3-4B      & 0.811 & 0.680 & 0.464 & 0.349 & 0.407 & 0.525 & 0.375 & 0.232 & 0.357 \\
Qwen3-8B      & 0.967 & 0.839 & 0.721 & 0.629 & 0.495 & 0.397 & 0.344 & 0.370 & 0.333 \\
Qwen3-14B     & 1.000 & 0.946 & 0.823 & 0.841 & 0.790 & 0.701 & 0.455 & 0.320 & 0.343 \\
Qwen3-30B     & 0.973 & 0.851 & 0.634 & 0.565 & 0.479 & 0.437 & 0.373 & 0.425 & 0.409 \\
\addlinespace[6pt]

\rowcolor{mygreen1} 
\multicolumn{10}{@{}l}{\small\textit{Gemma Series}} \\
\addlinespace[2pt]
Gemma3-1B      & 0.552 & 0.463 & 0.403 & 0.433 & 0.382 & 0.346 & 0.347 & 0.321 & 0.333 \\
Gemma3-4B      & 0.917 & 0.727 & 0.600 & 0.575 & 0.555 & 0.418 & 0.360 & 0.314 & 0.500 \\
Gemma3-12B     & 0.967 & 0.877 & 0.806 & 0.813 & 0.615 & 0.519 & 0.492 & 0.442 & 0.409 \\
Gemma3-27B     & 0.933 & 0.923 & 0.902 & 0.836 & 0.697 & 0.495 & 0.508 & 0.419 & 0.405 \\
\addlinespace[6pt]

\rowcolor{othercolor} 
\multicolumn{10}{@{}l}{\small\textit{Other Models}} \\
\addlinespace[2pt]
GPT-4.1 Nano  & 0.950 & 0.809 & 0.784 & 0.593 & 0.514 & 0.384 & 0.411 & 0.377 & 0.356 \\
GPT-4.1 Mini  & 0.950 & 0.912 & 0.874 & 0.826 & 0.701 & 0.704 & 0.591 & 0.535 & 0.435 \\
DeepSeek V3.1 & 0.953 & 0.863 & 0.873 & 0.779 & 0.718 & 0.596 & 0.481 & 0.473 & 0.469 \\

\bottomrule
\end{tabular}
\end{table*}

\section{Experimental Details}
\label{sec:appendix_experimental_details}

\noindent\textbf{Overview.}
This section provides detailed experimental settings and extended analyses to ensure reproducibility and support the main findings.
We describe the evaluation protocol and implementation details (\S\ref{sec:appendix_eval_impl}), report full results across \emph{Model $\times$ Complexity Bins} (\S\ref{app:full_results}), analyze the necessity of curriculum design (\S\ref{app:curriculum_necessity}), include an analysis of the FOLIO CoT anomaly (\S\ref{app:folio_analysis}), and present diagnostic analyses of the complexity metric and logical phase transitions (\S\ref{sec:appendix_lpt_analysis}, \S\ref{sec:appendix_metric_diag}).
We further examine potential confounding factors, including complexity–premise count relations (\S\ref{app:complexity-vs-length}) and reasoning effort (\S\ref{sec:appendix_reasoning_effort}).

\subsection{Evaluation and Implementation Details}
\label{sec:appendix_eval_impl}

\noindent\textbf{Evaluation and answer extraction.}
All models are prompted to output a JSON-formatted response containing the reasoning trace and final answer.
When malformed outputs occur, we apply a robust post-processing strategy to recover predictions by prioritizing explicit final-answer fields when available, and otherwise extracting the answer from the trailing completion using heuristic patterns.

\noindent\textbf{Quantized evaluation for fair comparison.}
Due to computational constraints, all open-source models used in our evaluation are tested under \texttt{llama.cpp} Q4 quantization.
To ensure fair comparison, we quantize our fine-tuned checkpoint to the same Q4 setting via \texttt{llama.cpp} before reporting results.

\noindent\textbf{Training setting.}
We fine-tune Qwen2.5-7B-Instruct using parameter-efficient LoRA on a single Tesla V100S-PCIE-32GB GPU.
Training is performed for one epoch with a learning rate of $1\times10^{-4}$ and an effective batch size of $48$, implemented with a per-device batch size of $1$ and \texttt{gradient\_accumulation\_steps}$=48$.
For LoRA, we set the rank to $r=8$ and the scaling parameter to $\alpha=16$.

\subsection{Full Results}
\label{app:full_results}

This section extends the analysis beyond the models reported in the main text.
The supplementary phase-transition curves shown in Figure~\ref{fig:Supplementary_LPT} include additional model families and sizes, providing complementary evidence for the observed logical phase-transition behavior across contemporary LLMs.

\noindent\textbf{Model-by-Bin Accuracy Table.}
For detailed numerical comparison, Table~\ref{tab:full_results_model_by_bins} reports the accuracy of each tested model across the nine complexity bins.
This tabular view enables precise inspection of performance at specific complexity levels and facilitates the identification of patterns that may not be immediately visible from the phase-transition curves alone.

\noindent\textbf{Transition patterns vary within a model family.}
Even within the same model family, both the \emph{number} of detected critical intervals and the \emph{onset} of logical phase transitions can vary substantially across parameter scales.
The Gemma series illustrates this variability: its 4B and 12B models exhibit two critical intervals, whereas the 1B and 27B models show only one.
This observation indicates that transition dynamics are not governed by a simple monotonic function of model size.

\noindent\textbf{Scaling does not guarantee robustness in high-complexity regimes.}
Increasing parameter count does not necessarily lead to higher overall accuracy or delayed phase transitions.
For example, Qwen2.5-14B achieves higher overall accuracy (70.2\%) than its 32B counterpart (68.3\%) and consistently outperforms it in the mid-to-high complexity range (Bin~5--Bin~8).
This suggests that scaling alone is insufficient for handling high-LoCM instances, and that phase-transition behavior may instead be influenced by factors such as pretraining data composition, instruction tuning, and optimization dynamics.

\noindent\textbf{Scaling consistently benefits low-complexity performance.}
In contrast to the non-monotonic behavior observed at higher LoCMs, scaling yields stable improvements in the low-complexity regime.
Moving from 1--4B to 12--14B models results in substantial and consistent gains in Bin~1--Bin~3.
These results indicate that while scaling effectively improves performance on elementary instances, it exhibits diminishing returns or instability when models approach the complexity threshold.

\subsection{Necessity of Curriculum Training}
\label{app:curriculum_necessity}
\begin{figure}[t]
    \centering
    \includegraphics[width=\linewidth]{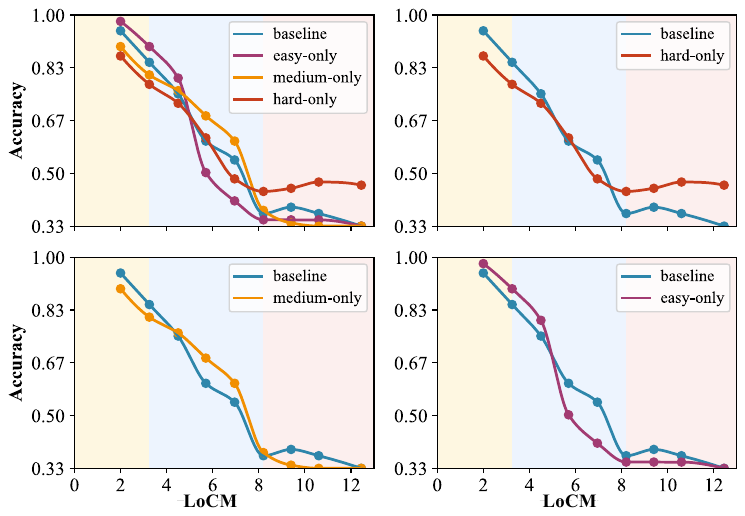}
    \caption{Effect of training on isolated complexity regimes.}
    \label{fig:curriculum_necessity}
\end{figure}

\textbf{Single-regime training is insufficient for robust reasoning across logical complexity levels.}
To examine this, we conduct controlled fine-tuning experiments in which training is restricted to a single complexity interval. As shown in Figure~\ref{fig:curriculum_necessity}, models trained exclusively on high-complexity instances (\textit{hard-only}) achieve better accuracy in the high-complexity region but suffer substantial degradation on low- and medium-complexity samples, indicating catastrophic forgetting of reasoning patterns associated with simpler regimes. By contrast, training restricted to low- or medium-complexity instances does not improve performance on high-complexity samples and can even further degrade that regime. Taken together, these results show that single-regime training cannot sustain consistent reasoning performance across complexity levels, motivating the use of complexity-aware curriculum training.

\subsection{Analysis of the FOLIO CoT Anomaly}
\label{app:folio_analysis}
As shown in Table~\ref{tab4:compare_with_baselines}, FOLIO is the main exception to the otherwise consistent gains of NSCT under CoT prompting. We do not view this as evidence of instability specific to NSCT, since similar drops on FOLIO also appear for several models fine-tuned on formal reasoning datasets, including the ProntoQA-tuned, ProofWriter-tuned, and FOLIO-tuned variants. This suggests that the anomaly more likely reflects a broader mismatch between formal-logic-oriented tuning and the reasoning demands of FOLIO.

Although FOLIO is annotated with first-order logic, it places heavier demands on natural-language interpretation and commonsense grounding than the more structurally controlled benchmarks. In contrast, NSCT is primarily designed to strengthen formal validity under increasing logical complexity. We therefore interpret the FOLIO result as a boundary case of the current framework rather than a contradiction of the main findings, and view it as a motivation for incorporating more commonsense-sensitive reasoning data in future work.

\subsection{Logical Phase Transition Analysis}
\label{sec:appendix_lpt_analysis}

\begin{figure}[t]
    \centering
    \includegraphics[width=1.0\linewidth]{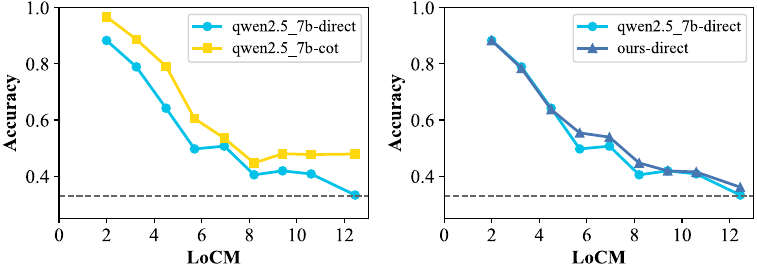}
    \caption{Logical phase-transition curves under different interventions. \textbf{Left:} Direct vs.\ CoT prompting on the base model. \textbf{Right:} base vs.\ fine-tuned model under Direct prompting.}
    \label{fig:phase_compare}
\end{figure}

In this section, we further probe the logical phase transition phenomenon by examining whether commonly used interventions—Supervised Fine-Tuning (SFT), prompting strategies, and model scaling—can shift or delay the collapse region.
Figure~\ref{fig:phase_compare} summarizes the resulting comparisons.

\noindent\textbf{Observation 1: Supervised Fine-Tuning yields local accuracy gains without substantially delaying the phase transition.}
As shown in the right panel of Figure~\ref{fig:phase_compare}, SFT improves performance across several complexity bins, while the sharp degradation still occurs within approximately the same LoCM range.
This indicates that although SFT better aligns the model with the task format and reduces pre-threshold errors, it does not significantly shift the critical interval itself.
In effect, fine-tuning primarily induces a \emph{vertical lift} in accuracy rather than a \emph{horizontal extension} of the complexity horizon.

\noindent\textbf{Observation 2: Advanced prompting strategies increase accuracy but do not extend the complexity horizon.}
The left panel of Figure~\ref{fig:phase_compare} shows that switching from Direct to CoT prompting consistently improves accuracy across bins, while the transition onset remains stable.
This suggests that CoT acts as an inference-time facilitator, enabling the model to better utilize its existing capacity, but leaving the high-complexity failure mode intact once logical difficulty exceeds that capacity.
Prompting therefore raises the overall accuracy profile without reliably shifting the phase-transition boundary.

\noindent\textbf{Observation 3: Scaling strengthens low-complexity robustness but does not eliminate high-complexity collapse.}
Referring to the scaling analysis in Figure~\ref{fig:Supplementary_LPT}, increasing model size consistently stabilizes performance in low-LoCM regimes, forming a more robust pre-threshold plateau.
However, the phase-transition behavior persists: even larger models eventually experience a rapid collapse once the complexity threshold is crossed.
This pattern suggests that scaling improves performance on simple instances but does not prevent collapse under high logical complexity.

\begin{figure}[htbp]
    \centering
    \includegraphics[width=0.95\linewidth]{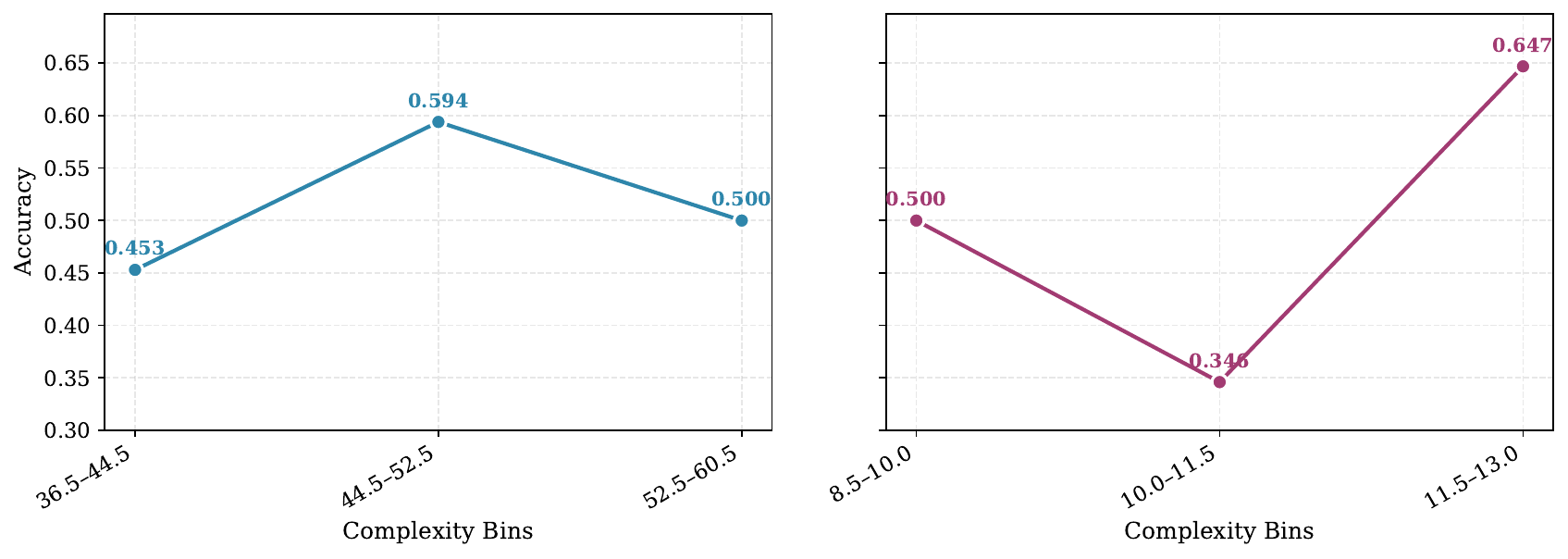}%
    \caption{Accuracy vs.\ single-operator complexity proxies (Left: $\land/\lor$; Right: $\forall/\exists$). }
    \label{fig:single-op-perf}
\end{figure}

\subsection{LoCM Diagnostics and Calibration}
\label{sec:appendix_metric_diag}
\noindent\textbf{Design Principle: Structural Orthogonality and Expressive Sufficiency.}
We construct the operator basis $\mathcal{O}=\{\lnot, \land, \lor, \rightarrow, \leftrightarrow, \oplus, \forall, \exists\}$
not merely as a syntactic collection, but as a set of \emph{orthogonal reasoning primitives} representing distinct cognitive operations in symbolic deduction.
This design is inspired by Russell's logical atomism, which views valid reasoning as discovering a finite set of irreducible logical forms whose truth-functional compositions constitute the "skeleton of the world."
While certain connectives are mathematically redundant in propositional logic, they introduce unique inferential patterns and computational burdens for LLMs---such as branching case-splits for $\oplus$ or variable binding for $\forall$---which are not captured by coarse complexity proxies.
Consequently, our basis is \emph{expressively sufficient} to cover the first-order semantics in our benchmarks while remaining \emph{structurally minimal}.
To account for global inference costs, we augment LoCM with a weighted hop term $h(\phi)$ (weighted by $\gamma=2$).
This formulation allows LoCM to distinguish between \emph{deep-but-narrow} and
\emph{shallow-but-dense} reasoning chains, capturing diverse empirical failure modes that surface despite similar local structures.

\noindent\textbf{The final weights reflect both intuition and validation-set calibration.}
We assign operator weights through a controlled grid search on a held-out validation set, using Pearson correlation between LoCM and model accuracy as a selection criterion.The resulting weights are summarized in Table~\ref{tab:LoCM_weights}.Importantly, the selected configuration aligns with intuitive differences in reasoning burden across operators.Exclusive OR ($\oplus$) typically induces higher branching factors and explicit case splits, motivating its larger weight, while conditionals ($\rightarrow,\leftrightarrow$) often require multi-branch implication handling.Quantifiers and negation introduce moderate structural bookkeeping, such as variable binding, scope management, or polarity switching, whereas basic connectives serve as the baseline.Beyond correlation-based selection, the necessity of the full weighted formulation is independently supported by the operator-weight ablations in Table~\ref{tab:LoCM_weight_corr}, where removing or isolating any operator family consistently weakens the observed correlation.

\begin{table}[h] 
\centering
\small
\setlength{\tabcolsep}{7pt}
\renewcommand{\arraystretch}{1.0} 
\caption{Calibrated operator weights used in LoCM calculation. }
\label{tab:LoCM_weights}
\begin{tabular}{lcc}
\toprule
\textbf{Component} & \textbf{Operator(s)} & \textbf{Weight} \\
\midrule
Basic connectives & $\land,\lor$ & 1.0 \\
Quantifiers & $\forall,\exists$ & 2.0 \\
Negation & $\lnot$ & 2.0 \\
Hop term & $h(\phi)$ & 2.0 \\
Conditionals & $\rightarrow,\leftrightarrow$ & 3.0 \\
XOR & $\oplus$ & 3.5 \\
\bottomrule
\end{tabular}
\end{table}

\noindent\textbf{Multi-operator composition is necessary to avoid length-driven artifacts.}
Single-operator heuristics tend to conflate \emph{surface length} with \emph{logical difficulty}.
As shown in Figure~\ref{fig:single-op-perf}, accuracy curves derived from individual operator counts (e.g., quantifier-only metrics) often exhibit \textbf{non-monotonic fluctuations} and spurious spikes in higher-complexity bins.
Such artifacts arise because longer derivations mechanically accumulate specific operators without necessarily increasing the underlying reasoning challenge.
By jointly modeling the full operator set in Eq.~\ref{eq:LoCM}, the complete LoCM mitigates operator-specific proxy leakage and produces smoother, more monotonic degradation trends.
This behavior more reliably exposes the underlying phase-transition structure, rather than artifacts induced by length or operator frequency alone.

\noindent\textbf{The square-root transform stabilizes binning by compressing heavy-tailed scores.}
The raw score
\[
S(\phi)=\sum_{o\in\mathcal{O}} \omega(o)\,\mathrm{freq}(o,\phi)+\gamma h(\phi)
\]
exhibits a heavy-tailed distribution, in which a small fraction of deeply nested instances attain disproportionately large values.
Directly binning samples based on $S(\phi)$ would therefore concentrate most instances into a narrow range, reducing resolution around the transition region.
To address this issue, we apply a sub-linear monotone transformation, defining 
\[
\mathrm{LoCM}(\phi)=\sqrt{S(\phi)}
\]
,
which preserves ordinal relationships while compressing extreme values.
As shown in Table~3, this square-root transformation yields the strongest (most negative) correlation with accuracy among the Linear, Log, and Square variants considered, supporting its use as a stable operationalization of logical complexity.

\noindent\textbf{LoCM is invariant to semantics-preserving serialization changes.}
As a sanity check, we verify that LoCM reflects underlying \emph{logical structure} rather than surface-level formatting.
Specifically, the metric remains unchanged under semantics-preserving perturbations such as variable renaming, whitespace variation, and equivalent parenthesizations.
This confirms that LoCM captures operator topology and hop structure, rather than artifacts of serialization.

\begin{figure}[t]
    \centering
    \includegraphics[width=1.00\linewidth]{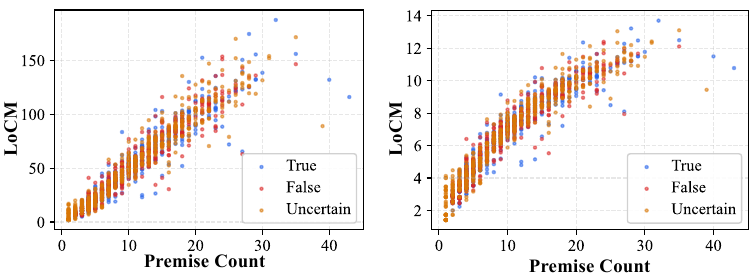}
    \caption{Premise count vs.\ complexity. \textbf{Left:} linear raw complexity score $S(\phi)$. \textbf{Right:} transformed metric $\mathrm{LoCM}(\phi)=\sqrt{S(\phi)}$.}
    \label{fig:premise-complexity}
\end{figure}

\subsection{LoCM vs.\ Premise Count}
\label{app:complexity-vs-length}
\noindent\textbf{LoCM correlates with premise count but is not reducible to length.}
A natural concern is that LoCM may act primarily as a proxy for input length, measured here by premise count.
Figure~\ref{fig:premise-complexity} plots premise count against (Left) the linear raw complexity score $S(\phi)$ and (Right) the transformed metric $\mathrm{LoCM}(\phi)=\sqrt{S(\phi)}$.
As expected, premise count and complexity exhibit a positive correlation, since longer derivations mechanically accumulate more logical operators.
However, the relationship is clearly non-deterministic: instances with similar premise counts can span a wide range of complexity values.
This dispersion indicates that LoCM captures substantial structural variation beyond length alone, reflecting differences in operator composition, nesting, and reasoning depth.
\begin{figure*}[!t]
    \centering
    \includegraphics[width=\linewidth]{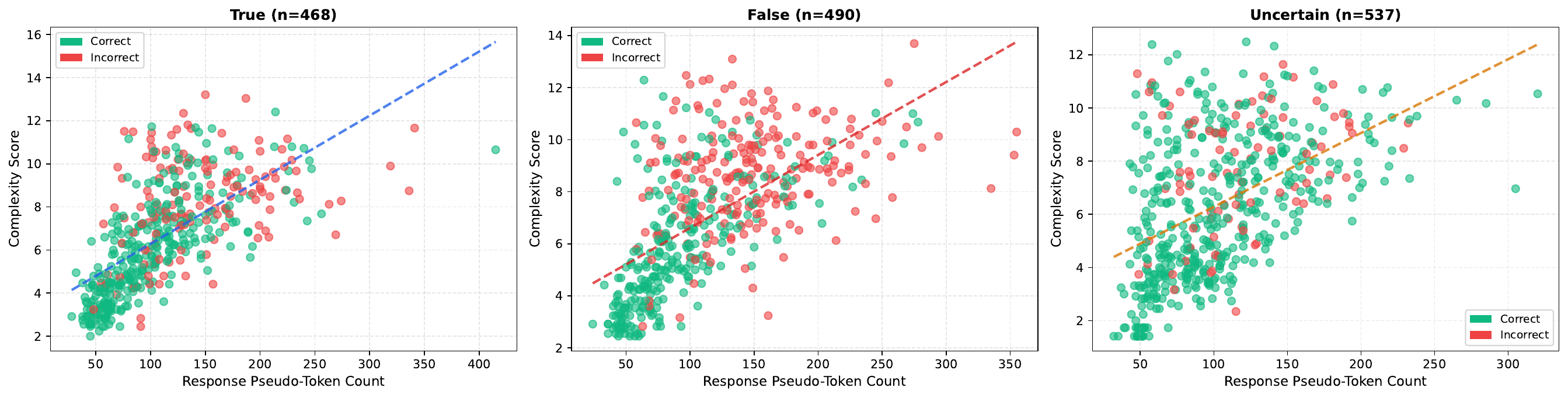}
    \caption{Completion length vs.\ LoCM, stratified by ground-truth label (Left: True; Middle: False; Right: Uncertain). Points are colored by correctness, and dashed lines denote linear fits.}
    \label{fig:complexity_effort_gt}
\end{figure*}

\noindent\textbf{The large vertical spread at fixed premise count indicates an independent structural signal.}
As shown in Figure~\ref{fig:premise-complexity}, for nearly any fixed premise-count value, instances span a wide range of raw complexity scores and corresponding LoCM values.
This vertical dispersion reflects differences between \emph{long-but-simple} structures (e.g., shallow chains dominated by $\land$/$\lor$) and \emph{structurally dense} reasoning patterns, such as deeper nesting, broader quantifier scopes, or higher-hop derivations, which premise count alone cannot distinguish.
The square-root transformation compresses extreme raw scores while preserving relative ordering, stabilizing the scale (Right) without collapsing this structural variation.

\noindent\textbf{Within narrow premise-count intervals, accuracy still decreases monotonically with LoCM.}
To assess whether LoCM captures performance variation beyond length, we perform a stratified control-variate analysis.
Specifically, we restrict instances to narrow premise-count intervals and examine accuracy trends across increasing LoCM bins.
As shown in Table~\ref{tab:control-variate}, accuracy consistently degrades as LoCM increases even when premise count is held approximately constant.
For instance, within the range $[1.0, 5.2]$, accuracy declines from $96.77\%$ to $85.39\%$ as LoCM increases, while in $[13.6, 17.8]$, it drops from $77.78\%$ to $49.57\%$.
These results indicate that LoCM captures structural reasoning difficulty beyond what can be explained by premise count alone.

\subsection{Reasoning Effort Analysis}
\label{sec:appendix_reasoning_effort}

\begin{table}[t]
\centering
\small
\caption{Accuracy vs.\ LoCM within fixed premise-count intervals}
\label{tab:control-variate}
\setlength{\tabcolsep}{8pt}
\renewcommand{\arraystretch}{1.1}
\begin{tabular}{c c r}
\toprule
\textbf{Premise-count interval} & \textbf{LoCM bin} & \textbf{Accuracy} \\
\midrule
\multirow{4}{*}{$[1.0,\,5.2]$}
 & $1.4$--$2.4$ & $96.77\%$ \\
 & $2.4$--$3.4$ & $95.30\%$ \\
 & $3.4$--$4.4$ & $93.08\%$ \\
 & $4.4$--$5.4$ & $85.39\%$ \\
\midrule
\multirow{4}{*}{$[13.6,\,17.8]$}
 & $5.1$--$6.3$ & $77.78\%$ \\
 & $6.3$--$7.4$ & $66.67\%$ \\
 & $7.4$--$8.5$ & $61.67\%$ \\
 & $8.5$--$9.7$ & $49.57\%$ \\
\bottomrule
\end{tabular}
\end{table}

\noindent\textbf{Observable decoding statistics provide a practical proxy for reasoning effort.}
We analyze reasoning effort using inference-time observable signals, primarily completion length (measured in pseudo-tokens), and relate them to answer accuracy and logical complexity (LoCM).
While response length is an imperfect proxy for internal computation, it provides a practical indicator of model behavior on challenging instances.

\noindent\textbf{Longer generations correlate with lower accuracy, signaling difficulty rather than improved reasoning.}
As shown in Table~\ref{tab:effort_accuracy}, accuracy decreases monotonically with increasing completion length, dropping from $94.66\%$ in the shortest bin to below $40\%$ in the tail.
Rather than reflecting better reasoning, extended generations more often indicate that the model has entered a difficult regime, expanding intermediate steps without reliably reaching a correct conclusion.

\begin{table}[t]
\centering
\small
\caption{Accuracy as a function of completion length. Accuracy decreases sharply as generation length increases.}
\label{tab:effort_accuracy}
\setlength{\tabcolsep}{10pt}
\renewcommand{\arraystretch}{1.1}
\begin{tabular}{l r}
\toprule
\textbf{token range} & \textbf{Accuracy} \\
\midrule
$24$--$63$   & $94.66\%$ \\
$63$--$102$  & $80.29\%$ \\
$102$--$141$ & $63.51\%$ \\
$141$--$180$ & $49.07\%$ \\
$180$--$220$ & $39.77\%$ \\
$220$--$259$ & $40.48\%$ \\
$259$--$415$ & $42.86\%$ \\
\bottomrule
\end{tabular}
\end{table}

\noindent\textbf{Effort increases with LoCM for definite labels, indicating length tracks structured difficulty in those regimes.}
We examine how completion length scales with LoCM, stratified by the ground-truth label (True / False / Uncertain).
As shown in Figure~\ref{fig:complexity_effort_gt}, the \texttt{True} and \texttt{False} subsets exhibit clear positive correlations between completion length and LoCM (Pearson $r=0.652$ and $r=0.618$, respectively), indicating that higher structural complexity tends to elicit longer generations when a definitive verdict exists.

\noindent\textbf{The Uncertain subset exhibits a high-variance, two-regime effort pattern.}
In contrast, the \texttt{Uncertain} subset shows substantially higher dispersion and a weaker correlation between completion length and LoCM (Pearson $r=0.485$).
The scatter reveals two regimes: (i) a short-response regime, where the model outputs \texttt{Uncertain} with minimal generation even at moderate-to-high LoCM, and (ii) a long-response regime, where extended derivations are produced with mixed correctness.
This bimodal pattern is consistent with heterogeneous decision behaviors, which we report descriptively without attributing it to a specific internal mechanism.

\section{Formal Setup and Notation}
\label{app:formal_setup}

\subsection{Notation and Basic Syntax}
\label{app:formal_setup:notation}

\noindent\textbf{Core symbols.}
We adopt standard first-order logic (FOL) notation. Variables ($x,y,z$) range over a domain of discourse, constants ($a,b,c$) denote specific objects, predicates ($P(\cdot),R(\cdot,\cdot)$) denote properties or relations, and functions ($f(\cdot),g(\cdot,\cdot)$) produce terms.
Logical connectives are drawn from the operator set
$\mathrm{OP}=\{\oplus,\ \vee,\ \wedge,\ \rightarrow,\ \leftrightarrow\}$, and quantifiers are $\forall$ and $\exists$.
Table~\ref{tab:fol_notation4} summarizes the key syntax elements used throughout the paper.

\noindent\textbf{Terms, formulas, and well-formedness.}
A \emph{term} is either a variable, a constant, or a function applied to terms.
An \emph{atomic formula} is a predicate applied to a sequence of terms.
A \emph{formula} is built from atomic formulas using connectives and quantifiers.
A formula is \emph{well-formed} (WFF) if it is syntactically valid under our grammar and can be evaluated as true or false under an interpretation.

\subsection{Well-Formedness and CFG Validation}
\label{app:formal_setup:cfg}

\noindent\textbf{CFG grammar.}
To ensure the well-formedness of all FOL expressions, we implement a symbolic parser using \texttt{nltk}~\citep{tool-nltk-2006nltk} and validate formulas with a context-free grammar (CFG):
\begin{center}\scriptsize
\begin{tabular}{rl}
S       & $\rightarrow$ F \ \textbar\ Q\ F \\
Q       & $\rightarrow$ \texttt{QUANT VAR} \ \textbar\ \texttt{QUANT VAR}\ Q \\
F       & $\rightarrow$ `$\lnot$' `(' F `)' \ \textbar\ `(' F `)' \ \textbar\ F\ OP\ F \ \textbar\ L \\
OP      & $\rightarrow$ `$\oplus$' \ \textbar\ `$\vee$' \ \textbar\ `$\wedge$' \ \textbar\ `$\rightarrow$' \ \textbar\ `$\leftrightarrow$' \\
L       & $\rightarrow$ `$\lnot$' PRED `(' TERMS `)' \ \textbar\ PRED `(' TERMS `)' \\
TERMS   & $\rightarrow$ TERM \ \textbar\ TERM `,' TERMS \\
TERM    & $\rightarrow$ CONST \ \textbar\ VAR \\
QUANT   & $\rightarrow$ `$\forall$' \ \textbar\ `$\exists$'
\end{tabular}
\end{center}

\noindent\textbf{Dynamic instantiation.}
Non-terminals \texttt{PRED}, \texttt{CONST}, and \texttt{VAR} are instantiated per example from the symbols appearing in the corresponding FOL context.
This allows the same CFG to validate diverse instances while enforcing a consistent syntactic structure.

\noindent\textbf{Syntactic validation.}
We apply CFG-based validation as a strict quality-control step before any downstream symbolic reasoning.
\textbf{Quantifier scope.} We require that each quantified variable is properly bound within its scope and appears as a valid \texttt{VAR} in the subsequent subformula.
\textbf{Predicate--argument form.} We enforce that every predicate application matches the pattern \texttt{PRED(TERMS)} with comma-separated terms, optionally preceded by $\lnot$.
\textbf{Operator placement.} We ensure that each binary connective in $\mathrm{OP}$ combines two valid subformulas and that parentheses yield an unambiguous parse under the CFG.

\subsection{NL-to-FOL Translation Rules}
\label{app:formal_setup:translation}

\noindent\textbf{Rule-based mapping.}
We translate NL premises into FOL using deterministic rules that map entities to constants, properties to unary predicates, and relations to $n$-ary predicates.
Logical operators and quantifiers are induced from explicit linguistic cues (e.g., ``every'', ``some'', ``if--then'', ``and/or'', ``not'').
Table~\ref{tab_app:fol_translation_rules} lists the translation rules used in our pipeline.

\noindent\textbf{Coreference and grounding.}
When NL contains pronouns or referential expressions (e.g., ``he'', ``this person''), we first resolve them to an existing entity mention in the local context and then reuse the corresponding constant in the translated formula to maintain symbol consistency.

\noindent\textbf{Negation patterns.}
We handle both local predicate negation (e.g., ``not honest'' $\mapsto \lnot Honest(x)$) and universal negation patterns (e.g., ``no citizen is armed'' $\mapsto \forall x\,(Citizen(x)\rightarrow \lnot Armed(x))$), ensuring that the resulting formulas remain WFF under the CFG in Appendix~\ref{app:formal_setup:cfg}.

\noindent\textbf{End-to-end check.}
Each translated formula is passed through CFG validation.
Only WFF expressions are retained for subsequent reasoning; malformed outputs are rejected and regenerated by the pipeline.

\section{Future Directions}
\label{app:future_directions}

\noindent\textbf{Complexity-aware curriculum learning for reinforcement learning.}
LoCM provides a structural foundation for curriculum design in logic-based reinforcement learning. By organizing instances according to LoCM, future training schedules can foster stable reasoning in low-complexity regimes before approaching the phase-transition boundary. Such a curriculum mitigates model collapse in difficult tasks by systematically bridging the gap between stable reasoning and the model's intrinsic complexity threshold

\noindent\textbf{Process-level supervision via verifiable symbolic reasoning chains.}
Standard RL for logic often struggles with sparse rewards and difficult credit assignment. Our Neuro-Symbolic alignment dataset addresses this by providing fully verifiable symbolic trajectories, enabling the development of Process Reward Models (PRMs). These models provide dense, step-wise feedback, replacing binary outcome-based rewards with structured verification signals that guide the model through complex reasoning paths.

\noindent\textbf{LoCM as a diagnostic tool for reasoning robustness.}
Building on the view of Russell~\cite{Principles-mathematics2020russell} and Wittgenstein~\cite{logico-philosophicus2023wittgenstein} that logic underlies the structure of the world, we treat LoCM not merely as a metric for symbolic tasks but as a general framework for measuring reasoning complexity across domains. Any domain with an explicit logical structure can be formalized into symbolic reasoning chains. By encoding domain-specific premises and inference rules in an LoCM-compatible form, researchers can identify model-specific complexity thresholds, positioning LoCM as a diagnostic tool for reasoning robustness and for analyzing how LLMs engage with the logical structure of knowledge.

\noindent\textbf{Phase transitions as operational capacity boundaries.} Our results suggest that logical phase transitions reflect intrinsic reasoning limits rather than artifacts of specific optimization. Future research should treat phase-transition behavior as an operational notion of reasoning capacity. A key direction is to investigate how different paradigms—such as reinforcement learning or architectural scaling—alter the sharpness or onset of this boundary without assuming it can be arbitrarily shifted.

\section{Full Prompts and Examples}
\label{sec:full-prompting}
Below are detailed prompts used in our evaluation.The prompts for constructing our datasets can be found in ProverGen~\citep{Benchmark-ProverQA-ICLR-2025large}.

\begin{table*}[h!]\scriptsize

\centering

\renewcommand{\arraystretch}{1.3}

\caption{Key Syntax Elements in First-Order Logic}

\label{tab:fol_notation4}

\begin{tabular}{p{2.5cm}p{3cm}p{7cm}}

\toprule

\textbf{Name} & \textbf{FOL Notation} & \textbf{Explanation} \\

\midrule

\textbf{Variable} & $x,\ y,\ z$ & Placeholder symbols representing arbitrary elements in the domain of discourse. \\

\textbf{Constant} & $a,\ b,\ c$ & Refer to specific, fixed objects in the domain. \\

\textbf{Operators (OP)} &  $\{$$\oplus$, $\vee$, $\wedge$, $\rightarrow$, $\leftrightarrow$$\}$ & 

Defines the set of logical connectives used to combine or relate propositions, including exclusive or, or, and, implication, and biconditional. Used in building compound formulas. \\

\midrule

\textbf{Function} & $f(x),\ g(x,y)$ & Maps input objects to an output object; returns a term. \\

\textbf{Predicate} & $P(x),\ R(x,y)$ & Express properties or relations; returns true or false. \\

\midrule

\textbf{Negation} & $\lnot P(x)$ & Logical NOT: $P(x)$ is not true. \\

\textbf{Conjunction} & $P(x) \land Q(x)$ & Logical AND: both $P(x)$ and $Q(x)$ must be true. \\

\textbf{Disjunction} & $P(x) \lor Q(x)$ & Logical OR: at least one of $P(x)$ or $Q(x)$ must be true. \\

\textbf{Implication} & $P(x) \rightarrow Q(x)$ & Logical implication: if $P(x)$ is true, then $Q(x)$ must be true. \\

\textbf{Biconditional} & $P(x) \leftrightarrow Q(x)$ & Logical equivalence: $P(x)$ and $Q(x)$ are true or false together. \\

\midrule

\textbf{Universal Quantifier} & $\forall x \; P(x)$ & “For all $x$, $P(x)$ is true” — generalization. \\

\textbf{Existential Quantifier} & $\exists x \; P(x)$ & “There exists $x$ such that $P(x)$ is true” — existential claim. \\

\midrule

\textbf{Term} & $x$, $a$, $f(a,x)$ & The basic expressions referring to objects (variables, constants, or functions). \\

\textbf{Atomic Formula} & $P(a,x)$ & A predicate applied to terms — indivisible logical unit. \\

\textbf{Complex Formula} & $\forall x (P(x) \rightarrow Q(f(x)))$ & A formula built from atoms using connectives and quantifiers. \\

\textbf{WFF (Well-formed)} & — & A syntactically valid FOL formula interpretable as true or false. \\

\bottomrule

\end{tabular}

\end{table*}
\begin{table*}[h!]\scriptsize
\centering
\renewcommand{\arraystretch}{1.25}
\caption{Natural Language to First-Order Logic (FOL) Translation Rules}
\label{tab_app:fol_translation_rules}
\begin{tabular}{p{2.8cm}p{4.6cm}p{7.4cm}}
\toprule
\textbf{Linguistic Element} & \textbf{NL Example} & \textbf{FOL Translation Rule and Output} \\
\midrule

\textbf{Entities / Objects} 
& ``Moriarty is a cat.'' 
& Nouns map to constants; adjectival properties map to unary predicates: \\
& & $Cat(Moriarty)$ \\

\textbf{Attributes / Properties} 
& ``Moriarty is fluffy.'' 
& Adjectives become unary predicates: \\
& & $Fluffy(Moriarty)$ \\

\textbf{Binary Relations} 
& ``Moriarty comforts a customer.'' 
& Verb relations become binary predicates: \\
& & $Comforts(Moriarty, cust)$ \\

\textbf{$n$-ary Relations} 
& ``Colt contributes to conservation.'' 
& Multi-argument verbs become $n$-ary predicates: \\
& & $Contributes(Colt, cons)$ \\

\midrule

\textbf{Universal Quantification} 
& ``All playful cats are fluffy.'' 
& Quantified NP induces $\forall$: \\
& & $\forall x\,(Playful(x) \land Cat(x) \rightarrow Fluffy(x))$ \\

\textbf{Existential Quantification} 
& ``Some cats are playful.''
& Existential statements induce $\exists$: \\
& & $\exists x\,(Cat(x) \land Playful(x))$ \\

\midrule

\textbf{Implication} 
& ``If a cat is playful, then it loves attention.'' 
& Conditionals map to implication: \\
& & $Playful(x) \rightarrow LovesAttn(x)$ \\

\textbf{Conjunction} 
& ``Colt is dedicated and makes discoveries.'' 
& Conjunction markers map to $\land$: \\
& & $Dedicated(Colt) \land Discovers(Colt)$ \\

\textbf{Disjunction} 
& ``A cat is either warm or calming.'' 
& Inclusive disjunction: \\
& & $Warm(x) \lor Calming(x)$ \\

\textbf{Exclusive-Or} 
& ``A cat is either warm or calming, but not both.'' 
& Exclusive contrast: \\
& & $Warm(x) \oplus Calming(x)$ \\

\midrule

\textbf{Negation} 
& ``Moriarty is not aggressive.'' 
& Negated predicates: \\
& & $\lnot Aggressive(Moriarty)$ \\

\textbf{Universal Negation} 
& ``No cat is aggressive.'' 
& ``No X is Y'' $\rightarrow$ universal implication with negation: \\
& & $\forall x\,(Cat(x) \rightarrow \lnot Aggressive(x))$ \\

\midrule

\textbf{Atomic Reasoning Step} 
& ``If a cat is fluffy and warm, then it comforts people.'' 
& Multi-premise reasoning: \\
& & $(Fluffy(x) \land Warm(x)) \rightarrow Comforts(x, ppl)$ \\

\textbf{Multi-step Chain} 
& ``If a cat is fluffy, then it is warm; if it is warm, then it comforts people.'' 
& Sequential implications: \\
& & $Fluffy(x) \rightarrow Warm(x),\ Warm(x) \rightarrow Comforts(x, ppl)$ \\

\midrule

\textbf{Coreference Resolution} 
& ``He is fluffy.'' (referring to Moriarty) 
& Resolve entity first: \\
& & $Fluffy(Moriarty)$ \\

\textbf{Role Assignment} 
& ``Colt studies plants.'' 
& Action roles become predicates: \\
& & $Studies(Colt, plants)$ \\

\bottomrule
\end{tabular}
\end{table*}

\clearpage
\tcbset{
     colback=blue!5,
     colframe=blue!75,
     fonttitle=\bfseries,
     width=\textwidth,
     boxrule=0.5mm,
     arc=4mm,
     title=Naive Prompting,
     breakable,
     enhanced,
     before skip=10pt,
     after skip=10pt
}

\begin{tcolorbox}
System:

Given a problem statement as contexts, the task is to answer a logical reasoning question.
Your answer should be in JSON format with key: answer.

Context:

has\_dense\_fur(Wynter):::Wynter has dense fur.

is\_playful(Wynter):::Wynter is playful.

has\_dense\_fur(Wynter) $\rightarrow$ (is\_playful(Wynter) $\oplus$ soft\_fur(Wynter)):::If Wynter has dense fur, then she is either playful or has soft fur, but not both.

Question:
Based on the above information, is the following statement true, false, or uncertain? 

Wynter is not beloved by others:::$\neg$beloved\_by\_others(Wynter)

Options:

A) True

B) False

C) Uncertain

The correct option is: \{

"answer": "C"

\}

-----------------

Other Examples

-----------------

Context:

[[CONTEXT]]

Question:

[[QUESTION]]

Options:

[[OPTIONS]]

The correct option is:

\end{tcolorbox}

\tcbset{
     colback=blue!5,
     colframe=blue!75,
     fonttitle=\bfseries,
     width=\textwidth,
     boxrule=0.5mm,
     arc=4mm,
     title=CoT Prompting,
     breakable,
     enhanced,
     before skip=10pt,
     after skip=10pt
}

\begin{tcolorbox}
System:

Given a problem statement as contexts, the task is to answer a logical reasoning question.
Your answer should be in JSON format with keys: reasoning, answer.

Context:

chooses\_words\_carefully(Peyton):::Peyton chooses words carefully.

listens\_attentively(Peyton):::Peyton listens attentively.

$\forall x\;((chooses\_words\_carefully(x) \lor listens\_attentively(x)) \rightarrow reserved\_speaker(x))$:::If a person chooses words carefully or listens attentively, then they are a reserved speaker.

\resizebox{\linewidth}{!}{$
\forall x\;((reserved\_speaker(x) \lor thoughtful\_listener(x)) \rightarrow builds\_meaningful\_connections(x))
$}
:::For all humans, if they are reserved speakers or thoughtful listeners, then they build meaningful connections.

Question:
Based on the above information, is the following statement true, false, or uncertain? 

Peyton does not build meaningful connections:::$\neg$builds\_meaningful\_connections(Peyton)

Options:

A) True

B) False

C) Uncertain

The correct option is: 
\end{tcolorbox}
\clearpage
\tcbset{
     colback=blue!5,
     colframe=blue!75,
     fonttitle=\bfseries,
     width=\textwidth,
     boxrule=0.5mm,
     arc=4mm,
     title=,
     breakable,
     enhanced,
     before skip=10pt,
     after skip=10pt
}
\begin{tcolorbox}
\{

"reasoning":

"fact1: chooses\_words\_carefully(Peyton):::Peyton chooses words carefully.

fact2: listens\_attentively(Peyton):::Peyton listens attentively.

rule:

\resizebox{\linewidth}{!}{$\forall x\;((chooses\_words\_carefully(x) \lor listens\_attentively(x)) \rightarrow reserved\_speaker(x))$}

:::If a person chooses words carefully or listens attentively, then they are a reserved speaker.

conclusion: reserved\_speaker(Peyton):::Peyton is a reserved speaker.

fact1: reserved\_speaker(Peyton):::Peyton is a reserved speaker.

rule:

\resizebox{\linewidth}{!}{$
\forall x\;((reserved\_speaker(x) \lor thoughtful\_listener(x)) \rightarrow builds\_meaningful\_connections(x))
$}

:::For all humans, if they are reserved speakers or thoughtful listeners, then they build meaningful connections.

conclusion: builds\_meaningful\_connections(Peyton):::Peyton builds meaningful connections.

Therefore, it is false that Peyton does not build meaningful connections. The correct option is: B.",

"answer": "B"

\}

-----------------

Other Examples

-----------------

Context:

[[CONTEXT]]

Question:

[[QUESTION]]

Options:

[[OPTIONS]]

The correct option is:
\end{tcolorbox}
\clearpage

 \lstset{
    backgroundcolor=\color{backcolour},   
    basicstyle=\ttfamily\scriptsize,
    breaklines=true,
    breakindent=20pt,
    keepspaces=true,
    columns=flexible,
    escapeinside={(*@}{@*)},
    frame=single,
    rulecolor=\color{gray!30},
    frameround=tttt
}

\begin{figure*}[t!]
\centering
\begin{lstlisting}
(*@\color{codepurple}{\textbf{Context \& Symbolic Mapping}}@*):
(*@\color{codegray}{Let $K := \text{Kaizen}$.}@*)

NL: Kaizen does not inspire cooperation.
FOL: (*@\color{codeblue}{$\neg\,\mathrm{inspires\_cooperation}(K)$}@*)
NL: Kaizen does not encourage learning.
FOL: (*@\color{codeblue}{$\neg\,\mathrm{encourages\_learning}(K)$}@*)
NL: If a philosophy leads to progress, then it inspires cooperation and encourages learning.
FOL: (*@\color{codeblue}{$\forall x\,\big(\mathrm{leads\_to\_progress}(x)\rightarrow(\mathrm{inspires\_cooperation}(x)\land \mathrm{encourages\_learning}(x))\big)$}@*)
NL: Kaizen's philosophy is transformative.
FOL: (*@\color{codeblue}{$\mathrm{is\_transformative}(K)$}@*)
NL: Kaizen challenges the status quo.
FOL: (*@\color{codeblue}{$\mathrm{challenges\_status\_quo}(K)$}@*)
NL: If Kaizen's philosophy is transformative, then it either challenges the status quo or embraces small changes, but not both.
FOL: (*@\color{codeblue}{$\mathrm{is\_transformative}(K)\rightarrow\big(\mathrm{challenges\_status\_quo}(K)\oplus \mathrm{embraces\_small\_changes}(K)\big)$}@*)
NL: Any philosophy that embraces small changes or inspires continuous change leads to progress.
FOL: (*@\color{codeblue}{$\forall x\,\big((\mathrm{embraces\_small\_changes}(x)\lor \mathrm{inspires\_continuous\_change}(x))\rightarrow \mathrm{leads\_to\_progress}(x)\big)$}@*)
NL: Kaizen's philosophy either inspires continuous change or supports growth, but not both.
FOL: (*@\color{codeblue}{$\mathrm{inspires\_continuous\_change}(K)\oplus \mathrm{supports\_growth}(K)$}@*)
NL: Kaizen advocates equality.
FOL: (*@\color{codeblue}{$\mathrm{advocates\_equality}(K)$}@*)
NL: If Kaizen's philosophy supports growth, then it advocates equality and encourages fair sharing.
FOL: (*@\color{codeblue}{$\mathrm{supports\_growth}(K)\rightarrow\big(\mathrm{advocates\_equality}(K)\land \mathrm{encourages\_fair\_sharing}(K)\big)$}@*)
NL: For any philosophy, it either encourages fair sharing or fosters social balance, but not both.
FOL: (*@\color{codeblue}{$\forall x\,\big(\mathrm{encourages\_fair\_sharing}(x)\oplus \mathrm{fosters\_social\_balance}(x)\big)$}@*)
NL: If Kaizen's philosophy promotes redistribution, then it fosters social balance.
FOL: (*@\color{codeblue}{$\mathrm{promotes\_redistribution}(K)\rightarrow \mathrm{fosters\_social\_balance}(K)$}@*)

(*@\color{codepurple}{\textbf{Question}}@*): 
NL: Based on the above information, is the following statement true, false, or uncertain?
    Kaizen's philosophy promotes redistribution.
Target FOL: (*@\color{codeblue}{$\mathrm{promotes\_redistribution}(K)$}@*)

(*@\color{codepurple}{\textbf{----------------------------------------Reasoning Trace (Dual-Track)----------------------------------------}}@*)

(*@\color{codegray}{\textbf{--Step 1--}}@*)
NL: If leading to progress would force inspiring cooperation and encouraging learning, then failing either one rules out leading to progress. Since Kaizen does not inspire cooperation (and also does not encourage learning), Kaizen does not lead to progress.
FOL: (*@\color{codeblue}{$\neg\,\mathrm{inspires\_cooperation}(K),\;\forall x(\mathrm{leads\_to\_progress}(x)\rightarrow(\mathrm{inspires\_cooperation}(x)\land \mathrm{encourages\_learning}(x)))\;\vdash\;\neg\,\mathrm{leads\_to\_progress}(K)$}@*)

(*@\color{codegray}{\textbf{--Step 2--}}@*)
NL: Being transformative enforces an exclusive choice: challenge the status quo or embrace small changes. Since Kaizen challenges the status quo, it cannot embrace small changes.
FOL: (*@\color{codeblue}{$\mathrm{is\_transformative}(K),\;\mathrm{challenges\_status\_quo}(K),\;\mathrm{is\_transformative}(K)\rightarrow(\mathrm{challenges\_status\_quo}(K)\oplus \mathrm{embraces\_small\_changes}(K))\;\vdash\;\neg\,\mathrm{embraces\_small\_changes}(K)$}@*)

(*@\color{codegray}{\textbf{--Step 3--}}@*)
NL: Anything that embraces small changes or inspires continuous change would lead to progress. Since Kaizen does not lead to progress, it cannot inspire continuous change (given it also does not embrace small changes).
FOL: (*@\color{codeblue}{$\forall x((\mathrm{embraces\_small\_changes}(x)\lor \mathrm{inspires\_continuous\_change}(x))\rightarrow \mathrm{leads\_to\_progress}(x)),\;\neg\,\mathrm{leads\_to\_progress}(K)\;\vdash\;\neg\,\mathrm{inspires\_continuous\_change}(K)$}@*)

(*@\color{codegray}{\textbf{--Step 4--}}@*)
NL: Kaizen must be exactly one of: inspires continuous change or supports growth. Since it does not inspire continuous change, it supports growth.
FOL: (*@\color{codeblue}{$\neg\,\mathrm{inspires\_continuous\_change}(K),\;(\mathrm{inspires\_continuous\_change}(K)\oplus \mathrm{supports\_growth}(K))\;\vdash\;\mathrm{supports\_growth}(K)$}@*)

(*@\color{codegray}{\textbf{--Step 5--}}@*)
NL: Supporting growth implies advocating equality and encouraging fair sharing; therefore Kaizen encourages fair sharing. (The standalone fact that Kaizen advocates equality is consistent but not needed for the derivation.)
FOL: (*@\color{codeblue}{$\mathrm{supports\_growth}(K),\;\mathrm{supports\_growth}(K)\rightarrow(\mathrm{advocates\_equality}(K)\land \mathrm{encourages\_fair\_sharing}(K))\;\vdash\;\mathrm{encourages\_fair\_sharing}(K)$}@*)

(*@\color{codegray}{\textbf{--Step 6--}}@*)
NL: Fair sharing and social balance are mutually exclusive (XOR). Since Kaizen encourages fair sharing, it does not foster social balance.
FOL: (*@\color{codeblue}{$\mathrm{encourages\_fair\_sharing}(K),\;\forall x(\mathrm{encourages\_fair\_sharing}(x)\oplus \mathrm{fosters\_social\_balance}(x))\;\vdash\;\neg\,\mathrm{fosters\_social\_balance}(K)$}@*)

(*@\color{codegray}{\textbf{--Step 7--}}@*)
NL: Promoting redistribution would imply fostering social balance. Since Kaizen does not foster social balance, it does not promote redistribution.
FOL: (*@\color{codeblue}{$\neg\,\mathrm{fosters\_social\_balance}(K),\;(\mathrm{promotes\_redistribution}(K)\rightarrow \mathrm{fosters\_social\_balance}(K))\;\vdash\;\neg\,\mathrm{promotes\_redistribution}(K)$}@*)

(*@\color{codepurple}{\textbf{Answer}}@*):
Therefore, the statement is (*@\color{codered}{\textbf{False}}@*). The correct option is B.
\end{lstlisting}

\caption{Detailed case study of a high-complexity instance ($LoCM \approx 7.25$) from the NSA-LR dataset. The example demonstrates strict NL-FOL alignment across a 7-step reasoning chain involving diverse logical operators ($\neg, \rightarrow, \oplus, \forall$).}
\label{fig:kaizen_case_study}
\end{figure*}

\end{document}